\documentclass{article}

\usepackage{iclr2027_conference,times}

\iclrfinalcopy
\def\arxivpreprint{}

\usepackage[disable,textsize=tiny]{todonotes}

\usepackage[utf8]{inputenc} %
\usepackage[T1]{fontenc}    %
\usepackage{hyperref}       %
\usepackage{url}            %
\usepackage{booktabs}       %
\usepackage{amsfonts}       %
\usepackage{nicefrac}       %
\usepackage{microtype}      %
\usepackage{xcolor}         %

\usepackage{graphicx}
\graphicspath{{common/figures/}}
\usepackage{subcaption}
\usepackage{amsmath}
\usepackage{amssymb}
\usepackage{mathtools}        %
\usepackage{amsthm}
\usepackage[capitalize,noabbrev]{cleveref}

\theoremstyle{plain}
\newtheorem{theorem}{Theorem}[section]
\newtheorem{proposition}[theorem]{Proposition}

\theoremstyle{definition}

\theoremstyle{remark}

\let\originalleft\left
\let\originalright\right
\renewcommand{\left}{\mathopen{}\mathclose\bgroup\originalleft}
\renewcommand{\right}{\aftergroup\egroup\originalright}

\newcommand{\reals}{\mathbb{R}}
\newcommand{\R}{\reals}

\DeclareMathOperator{\E}{\mathbb{E}}
\DeclareMathOperator*{\UE}{\mathbb{E}}

\newcommand{\EI}{\mathrm{EI}}

\DeclareMathOperator{\Geo}{Geo}

\DeclareMathOperator{\Unif}{Unif}

\newcommand{\simiid}{\overset{\text{i.i.d.}}{\sim}}

\newcommand{\cX}{\mathcal{X}}
\newcommand{\cY}{\mathcal{Y}}

\newcommand{\llm}{\pi^{\scriptscriptstyle \text{LLM}}}

\usepackage{xparse}

\NewDocumentCommand{\profit}{m o o}{
  V_{#1\IfValueT{#2}{,#2}}%
  \IfValueT{#3}{^{#3}}
}

\makeatletter
\newtoks\apx@floats
\newcommand\apx@float[2]{\begingroup\@float{#1}[ht]#2\end@float\endgroup}
\newcommand\apx@hold[2]{%
  \global\apx@floats\expandafter{\the\apx@floats\apx@float{#1}{#2}}}
\newcommand\apx@drain{%
  \ifx\@deferlist\@empty\else\vbox{}\newpage\expandafter\apx@drain\fi}
\newcommand\apx@flush{%
  \edef\apx@tmp{\detokenize\expandafter{\the\apx@floats}}%
  \ifx\apx@tmp\@empty\else
    \par
    \global\@topnum\z@
    \begingroup\the\apx@floats\endgroup
    \global\apx@floats{}%
    \ifx\@deferlist\@empty\else\newpage\apx@drain\fi
  \fi}
\newcommand\AppendixFloatsBySection{%
  \let\apx@origsection\section
  \def\section{\apx@flush\apx@origsection}%
  \let\apx@origsubsection\subsection
  \def\subsection{\apx@flush\apx@origsubsection}%
  \let\apx@origfigure\figure \let\apx@origendfigure\endfigure
  \let\apx@origtable\table   \let\apx@origendtable\endtable
  \RenewDocumentEnvironment{figure}{o +b}{\apx@hold{figure}{##2}}{}%
  \RenewDocumentEnvironment{table}{o +b}{\apx@hold{table}{##2}}{}%
  \setcounter{topnumber}{20}\setcounter{totalnumber}{20}%
  \renewcommand\topfraction{1}\renewcommand\textfraction{0}}
\newcommand\EndAppendixFloats{%
  \apx@flush
  \let\section\apx@origsection
  \let\subsection\apx@origsubsection
  \let\figure\apx@origfigure \let\endfigure\apx@origendfigure
  \let\table\apx@origtable   \let\endtable\apx@origendtable}
\makeatother

\title{Latency-Aware Adaptive Best-of-$N$}

\author{%
  Alon Jacobson, Qian Xie \& Peter I.\ Frazier \\
  School of Operations Research and Information Engineering \\
  Cornell University \\
  Ithaca, NY 14853, USA \\
  \texttt{\{adj53,qx66,pf98\}@cornell.edu}
}

\begin{document}

\maketitle
\lhead{Preprint}

\begin{abstract}
Test-time compute (TTC) improves language model responses by spending additional computation at inference time, and the right amount of compute depends on the prompt. Adaptive TTC methods dynamically allocate compute based on signals of response quality such as observed rewards, often by stopping once the signal exceeds a fixed threshold; for Best-of-$N$ sampling, a widely used test-time compute method, with reward as the signal, this is the same as stopping once the expected reward improvement falls below a fixed cost. Adaptive TTC generally measures cost by the number of generations or tokens, which suits fixed offline workloads. But in online serving, latency is also a cost, and can vary significantly: it grows with server load and is affected by prompt and response lengths, so a method using a fixed threshold spends too little compute when generation is fast and too much when it is slow. We propose a two-stage adaptive Best-of-$N$ algorithm that draws an initial batch of responses, then uses the observed rewards and the time taken to generate and score them to decide whether to draw a second batch and at what size by comparing expected improvement in reward against predicted cost: it draws a second batch when the first batch's best reward is below a threshold that decreases with the first batch's latency, so it is less likely to draw one when the first was slow, and more likely when it was fast, with no need to explicitly measure server load. We map each reward to its value under the cumulative distribution function (CDF) of the prompt's reward distribution, so that the same tradeoff parameters can be reused across tasks without retuning; a regression fitted offline on other prompts predicts a distribution for the CDF value of the first batch's best reward, from which the expected improvement follows in closed form. We evaluate the method by simulation on a preference-based alignment benchmark, measuring cost as \emph{generations-equivalent cost} -- the number of generations plus the latency, converted to generations. The method reaches Best-of-$N$'s reward at lower generations-equivalent cost, and conditioning on the measured first-batch latency saves at least as much of that cost as fixed-threshold (latency-unaware) baselines, and saves more when latency is highly variable, when latency grows with batch size, and when the second batch's latency is highly predictable from the first's.
\end{abstract}

\section{Introduction}
\label{sec:intro}

Test-time compute (TTC) methods improve the quality of language model (LM) outputs by spending additional computation at inference time \citep{welleck2024metageneration,snell2024scaling}. A central challenge is to decide \emph{how much} compute to spend per prompt: too little produces lower-quality responses, while too much wastes compute. The right amount of compute depends on the prompt: easy prompts, such as simple fact lookups, need very little compute, while hard prompts, such as complex mathematics problems or problems with high response quality requirements, continue to benefit from additional compute.

A line of recent work proposes \emph{adaptive} test-time compute algorithms, which allocate less compute to easier prompts and more to harder ones.
Many of them work by using signals of response quality such as observed rewards, curtailing generation when additional response generations or thinking tokens are unlikely to improve the final response; a common way of doing this is by stopping generation once a quality signal, such as reward, exceeds a fixed threshold. Such policies are usually heuristics, setting the threshold ad-hoc or by optimizing on a fixed validation dataset.

These methods almost universally evaluate cost by the number of tokens or generations used, which is appropriate for fixed offline workloads. But test-time compute, notably in reasoning models, is also widely deployed in online serving \citep{xiang2026servegen,zhang2026fineserve}, where many requests are being processed concurrently and the latency it takes to serve each request needs to be taken into account in order to ensure consistent system throughput and overall user satisfaction. Furthermore, in these scenarios the latency varies significantly: it grows with system load, and is also affected by prompt and output lengths (\cref{sec:serving}).
A method that stops at a fixed threshold therefore spends too little compute when generation is fast (continuing would be cheap) and too much when it is slow (continuing is expensive).

We focus on \emph{Best-of-$N$ sampling}, a widely used TTC method: given a prompt, generate $N$ responses independently from the LM, score each with a reward model, and return the highest-scoring response (\cref{sec:lm-rm-bon}). One way to make Best-of-$N$ adaptive is to generate responses, or batches of responses, sequentially until the reward exceeds a threshold, or equivalently until the expected reward improvement from continuing falls below a fixed cost; when the reward distribution is known, the threshold is fixed and this threshold policy is optimal for a cost-penalized reward objective, which some adaptive Best-of-$N$ works \citep{kalayci2025optimal,wan2025beacon} explicitly model and optimize. Because the $N$ responses are independent, all $N$ can be generated in parallel if capacity allows, so generating one response at a time, although optimal for saving generations, is far from optimal when latency is taken into account.

\begin{figure}[t]
  \centering
  \begin{subfigure}[b]{0.555\textwidth}
    \includegraphics[width=\linewidth]{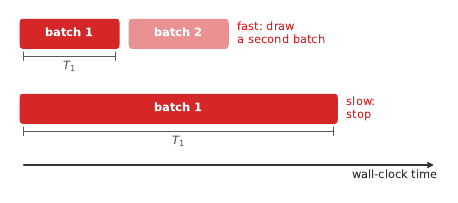}
    \caption{}
    \label{fig:overview_timeline}
  \end{subfigure}\hfill
  \begin{subfigure}[b]{0.44\textwidth}
    \includegraphics[width=\linewidth]{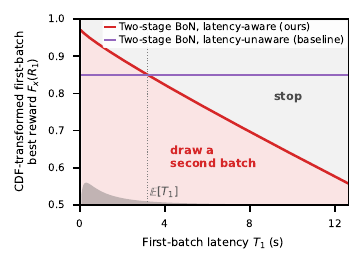}
    \caption{}
    \label{fig:overview_rule}
  \end{subfigure}
  \caption{(a) Our policy draws a second batch after a fast first batch and stops after a slow one. (b) It draws a second batch, of adaptively chosen size, when the first batch's CDF-transformed best reward $F_x(R_1)$ is below a threshold that falls with the first batch's latency $T_1$; the latency-unaware policy's threshold is fixed (\cref{apx:threshold}). Grey: the distribution of $T_1$. Settings (\cref{sec:experiments:setup}): $F_x$ known, $\beta = 0.5$, $c_{\mathrm{v}} = 2$, $\rho = 1$, $\lambda_{\text{gen}} = 10^{-3.5}$, $\lambda_{\text{lat}} = 10^{-1.5}$.}
  \label{fig:overview}
\end{figure}

We propose a \emph{two-stage} adaptive Best-of-$N$ policy, novel among adaptive Best-of-$N$ methods even disregarding latency awareness; we show that significant cost savings can be achieved even with this two-batch limitation, and argue that more batches would incur too much latency in online serving. We first generate a batch of $N_1$ responses and observe their rewards together with the latency $T_1$ taken to generate and score them. Next, we draw a second batch, and choose its size $N_2$, only when the expected reward improvement from generating the second batch exceeds its predicted cost -- which depends on both its latency $T_2$ and batch size $N_2$. The key insight is that $T_1$ is an informative predictor of $T_2$: whatever makes the first batch slow -- server load, scheduling, prompt length and response lengths -- largely persists over the short interval before the second batch is submitted. The policy is less likely to generate a second batch when generation is slow, whatever the cause, with no explicit load monitoring required; it is also a threshold-based stopping policy, where the threshold decreases with $T_1$ (\cref{fig:overview}). \citet{tang2026laser} also reduce a threshold when load is high, for chain-of-thought and based on measured load, but to our knowledge, no prior adaptive TTC method conditions its compute decisions on real-time measured latency.
We use a reward transformation based on the prompt-conditional cumulative distribution function (CDF). Because reward distributions have arbitrarily different scales across prompts, language models, and reward models, the transformation makes the tradeoff parameters scale-free: the same values can be reused across tasks without per-task retuning.
To predict the expected CDF-transformed reward improvement from the first-batch rewards, we combine simple parametric estimation with a regression model fitted offline on other prompts' rewards that predicts a Beta distribution for the CDF value of the first batch's maximum reward, yielding a closed-form expected improvement requiring no Monte Carlo estimation.
Empirically, measuring cost as \emph{generations-equivalent cost} -- the number of generations plus the latency, converted to generations -- our policy reaches Best-of-$N$'s reward at lower cost, more so with adaptive batch sizes, and conditioning on $T_1$ saves at least as much of that cost as fixed-threshold (latency-unaware) baselines, saving more when latency is highly variable, when it grows with batch size, and when the second batch's latency is highly correlated with -- and thus highly predictable from -- the first's.

\subsection{Related work}
\label{sec:related}

\paragraph{Adaptive test-time compute.}
A growing body of work adapts the compute spent on a request to its difficulty \citep{welleck2024metageneration,alomrani2025survey,snell2024scaling}, using signals of response quality such as observed rewards, verifier scores, or self-consistency \citep{wang2023selfconsistency}, and treats the cost of compute as fixed, ignoring the realized latency. Many stop once their signal exceeds a threshold fixed offline \citep{aggarwal2023let,li2024escape,yang2026dynamic,fu2025dynasor}.

\paragraph{Adaptive Best-of-$N$.}
Several recent methods are specifically tailored to Best-of-$N$. \citet{kalayci2025optimal} cast inference-time alignment as an infinite-horizon Pandora's Box problem \citep{weitzman1979optimal}, generating responses one at a time and stopping once a reward exceeds an estimated Gittins index. Their main experiments are sequential, impractical in parallel serving stacks, where doubling per-request latency to halve compute is rarely worthwhile. %
Similar to ours, they use a reward transformation that makes reward scales comparable across prompts: a heuristic ``acceptance probability'', from a Bradley--Terry surrogate, which is less principled than our CDF transformation (\cref{sec:cdf-transformation}). 
BEACON \citep{wan2025beacon} places a conjugate Normal prior on a prompt's rewards and stops when a Bayesian index falls below a fixed per-sample cost, in its parallel variant updating the posterior after each fixed-size batch; LAND \citep{oostermeijer2026land} instead fits a mixture of reward distributions on calibration prompts and stops when the expected improvement of another response falls below a cost set to meet a target average number of generations; and \citet{manvi2024adaptive} resample in geometrically growing batches until a fine-tuned self-evaluation's win-or-tie probability -- equivalently a predicted CDF value of the response's reward up to a tie tolerance, similar to our CDF transformation -- exceeds a fixed threshold. Each of these four draws responses or batches until its signal exceeds a threshold, as our latency-unaware baseline does, but is not capped at two batches; the threshold is fixed offline in \citeauthor{manvi2024adaptive}'s method, and in the other three is recomputed from a fixed cost and the updated reward distribution estimate.
These problems can be modeled as a batched Pandora's Box problem with a known per-batch setup cost as considered in \citet{dughmi2026batched}; with a known reward distribution and unlimited batches, the optimal policy draws batches of a fixed size until the maximum reward exceeds a fixed threshold. Capped at two batches, the optimal second batch size depends on the first batch (our adaptive batch size policy).

\emph{Speculative Rejection} \citep{sun2024fast} prunes low-reward responses mid-generation from extrapolated partial scores, and \emph{ST-BoN} \citep{wang2025stbon} keeps the generation whose early hidden states agree most with the others'. Both reject candidates before decoding completes rather than allocating from the rewards of completed ones; both could compose with our rule.

\citet{damani2024learning} make Best-of-$N$ allocation input-adaptive by training an auxiliary model to predict each prompt's marginal reward gain from additional samples, then distributing a fixed total budget across a batch of prompts by those predictions; the model must be retrained for each new domain, LM, decoding strategy, or budget. \emph{AdaBoN} \citep{raman2025abon} retains the input-adaptive idea but needs no auxiliary training: as we do, it draws two batches, sizing the second from the first batch's rewards. In both, however, adaptivity comes from \emph{amortizing} a budget across many prompts in a fixed offline workload; for a single prompt, the relevant case in online serving, AdaBoN always draws the second batch at the maximum budget, matching Best-of-$N$'s generations at twice the latency, and so is strictly dominated by it.

\paragraph{Load- and latency-aware test-time compute.}
\citet{fu2025dynasor} integrate adaptive test-time compute with an online serving stack. Their \emph{certaindex}, a reasoning-progress signal defined separately for each TTC algorithm -- self-consistency, Monte Carlo tree search (MCTS) \citep{hao2023reasoning,wan2024alphazero}, chain-of-thought, and Rebase \citep{wu2025inference} -- is the signal of a fixed-threshold stopping rule within a fixed compute cap, and their Dynasor system schedules the workloads on SGLang \citep{zheng2024sglang}. The threshold is set offline by validation-accuracy profiling, not derived from a quality--cost objective, so it must be re-tuned for each (algorithm, dataset) pair. Dynasor is load-aware in scheduling but not in stopping. For Best-of-$N$, their structure amounts to
our latency-unaware, fixed batch size variant (\cref{sec:algorithm}). \citet{huang2025latency} maximize, as we do, a quality measure (accuracy) minus penalties on token count and latency. Their adaptivity lies in a trained probe that predicts accuracy from the prompt embedding and strategy features, used to choose a TTC strategy and budget per query before generation rather than mid-generation; the costs are each strategy's average token count and wall-clock time, precomputed on the training set, so they do not respond to load at request time. The only method we know of whose compute choice does is LASER \citep{tang2026laser}, which ties the chain-of-thought early-exit threshold of DEER \citep{yang2026dynamic} to the number of active requests on an edge server through a hand-tuned map; ours instead follows from a reward--latency objective and responds to the request's own measured latency, with no access to server internals, and is applied to Best-of-$N$ rather than chain-of-thought.

\section{Preliminaries}
\label{sec:preliminaries}

\subsection{Language models, reward models, and Best-of-$N$ sampling}
\label{sec:lm-rm-bon}

Let $\cX$ denote the space of prompts and $\cY$ denote the space of responses. A \emph{language model} (LM) $\llm : \cX \to \Delta(\cY)$ is a mapping from prompts to distributions over responses. A \emph{reward model} is a function $r : \cX \times \cY \to \R$ that maps a prompt and a response to a real number, with the property that for a fixed prompt, responses with higher reward values are preferable to responses with lower reward values.
The reward model may capture human preferences for alignment \citep{ouyang2022training,stiennon2020summarize}, or measure correctness for math and coding tasks \citep{cobbe2021gsm8k}.
Given a prompt $x \in \cX$, an LM $\llm$, and a reward model $r$, denote by $D_{r,\llm(x)} \in \Delta(\R)$ the distribution over rewards $r(x,y)$ when $y \sim \llm(x)$. We will write $F_x$ for the CDF of $D_{r,\llm(x)}$ when the LM and reward model are fixed and clear from context.

\emph{Best-of-$N$ sampling} draws $N$ responses $y_1, \ldots, y_N \simiid \llm(x)$, evaluates rewards $r_i = r(x, y_i)$, and returns the response $y_{i^\star}$ with $i^\star \in \arg\max_{i \in [N]} r_i$. Additional generations have diminishing marginal returns \citep{raman2025abon}. Best-of-$N$ therefore induces a fundamental \emph{reward--compute tradeoff}: larger $N$ yields higher expected reward at the price of additional generation cost.

\subsection{LLM serving and response latency}
\label{sec:serving}

We consider Best-of-$N$ running on a serving system that hosts $\llm$ and $r$ and processes many concurrent requests, as in some deployed applications \citep{irvine2023rewarding,zeng2023let}; we take that stack as given. We measure cost by two quantities the client observes: the \emph{number of generations} drawn, a proxy for the client's token cost and the system's compute cost, and the \emph{response latency} $T$, the wall-clock time between submitting a request and receiving its scored outputs.

Unlike the number of generations, latency is not a function of the algorithm's compute demand alone. We submit the $N$ responses of a draw together, as a single \emph{batch}, and observe one wall-clock latency $T$ for the batch to be generated and scored. Serving systems interleave concurrent requests \citep{kwon2023vllm,yu2022orca}, so $T$ depends on the server state -- the other requests in flight -- as well as on $N$. Load raises latency in two ways: more sequences decoded together can lengthen each decoding step, raising the time per output token; and when the server is full, new requests wait to begin decoding, raising the time to first token \citep{mitzenmacher2025queueing}. Larger batches can take longer for three reasons: when the server cannot decode all $N$ responses at once, some wait to begin decoding, and latency approaches the sum of their decoding times rather than the longest; when they are decoded in parallel, the batch completes only when its longest response does, which is longer in expectation for larger $N$; and its $N$ sequences add to those decoded together, raising the time per output token, though negligibly when $N$ is small relative to the number of sequences the server is decoding. Our method allows latency to depend on batch size, and the experiments of \cref{sec:experiments} vary how much it does (\cref{sec:experiments:setup}).

We assume that requests from adaptive policies such as ours are a small share of the server's load, the rest being, for example, single-response or fixed Best-of-$N$ requests. Their stopping decisions then do not materially change the load that other clients face, so we may optimize a single client's objective rather than an equilibrium among clients; if most requests came from adaptive policies, their stopping decisions would jointly set the load. Beyond this, we treat the form of latency as unknown to the client and rely on only two structural facts: that one batch's latency is informative about that of later batches submitted to the same server in the near future, and that the first batch's latency is measured before the second batch must be decided.

\subsection{The CDF reward transformation}
\label{sec:cdf-transformation}
Preference reward models are trained via the Bradley-Terry objective, which captures only \emph{relative} preferences within a prompt; raw scores are therefore not comparable across prompts, LMs, or reward functions, making it unclear what it means to subtract a latency cost from a raw reward. The CDF transformation $F_x$ maps each reward to its prompt-conditional CDF value -- equivalently, the probability that a fresh draw from the LM would score lower, i.e., a \emph{win rate}. This yields two benefits: (i)~\emph{scale-free parameters} -- the same $\lambda$ applies across tasks without retuning, because $F_x(r)\in[0,1]$ regardless of reward scale; and (ii)~a \emph{closed-form expected improvement} -- since $F_x(r)\sim\Unif[0,1]$ for $r\sim D_{r,\llm(x)}$, the quantity $\EI(u,N_2)$ derived in \cref{sec:latency-aware-policy} has an analytic expression requiring no Monte Carlo estimation. For outcome-based reward models (e.g., math or coding accuracy), whose rewards are comparable across prompts, the CDF transform remains valid but optimizes per-prompt relative performance; the identity transformation, with $\EI$ computed by Monte Carlo, optimizes mean accuracy instead. Additionally, the closed form for $\EI(u, N)$ requires a continuous reward distribution $D_{r,\llm(x)}$, so that $F_x(r) \sim \Unif[0,1]$; for discrete rewards, such as binary pass/fail, $\EI$ must be computed from the discrete distribution.

\section{Latency-aware two-stage Best-of-$N$}
\label{sec:algorithm}
Assume we are given parameters $\lambda_{\text{gen}}, \, \lambda_{\text{lat}} \geq 0$ which control the tradeoff between obtaining a response with high final reward and incurring low generation cost and latency, and finite sets $\mathcal{Z}_1, \, \mathcal{Z}_2 \subseteq \mathbb{N}$ of batch sizes to choose from in the first and second batches (our proposed \emph{adaptive batch size} policy (\cref{sec:batch_size_variants}) sets $\mathcal{Z}_1=\mathcal{Z}_2=\mathcal{Z} \subseteq \mathbb{N}$).
Throughout, we apply the CDF transformation $F_x$ to rewards (\cref{sec:cdf-transformation}).

When a new prompt (request) $x$ is received, an initial batch of $N_1 \in \mathcal{Z}_1$ responses is requested at once, generated and scored with a reward model, resulting in rewards $r^{(1)}_1, \ldots, r^{(1)}_{N_1} \simiid D_{r,\llm(x)}$ and giving a maximum reward of $R_1 = \max_{1 \leq i \leq N_1} r^{(1)}_i$. Generating and scoring the batch is measured to take $T_1$ seconds and incurs a cost of $c_1 = \lambda_{\text{gen}} N_1 + \lambda_{\text{lat}} T_1$.

Next, the policy $\pi_1$ picks an action $a_1$ from the following:
\begin{itemize}
  \item \emph{Generate another batch} of a chosen size $a_1 = N_2 \in \mathcal{Z}_2$ of new responses, giving rewards $r^{(2)}_1, \ldots, r^{(2)}_{N_2} \simiid D_{r,\llm(x)}$ and a maximum reward of $R_2 = \max_{1 \leq i \leq N_2} r^{(2)}_i$ for the batch. Generating and scoring it is measured to take $T_2$ seconds and incurs a cost of $c_2 = \lambda_{\text{gen}} N_2 + \lambda_{\text{lat}} T_2$, and the highest-reward response is chosen with a final utility of $\max\{F_x(R_1), F_x(R_2)\}$.
  \item \emph{Stop generation} ($a_1 = \mathrm{stop}$) and output the response with highest reward observed so far, collecting a final utility of $F_x(R_1)$.
\end{itemize}
The action $a_1 = \pi_1(s_1)$ is chosen based on the \emph{state} $s_1 = (r^{(1)}_1, \ldots, r^{(1)}_{N_1}, T_1)$, which includes the rewards observed in the first batch and the latency the first batch took. We want to choose the policy that maximizes the profit $V_\pi$, defined as the expected net utility:
\begin{equation} \label{eq:value}
  V_\pi \coloneqq \E_\pi\left[
    \max_{1 \leq i \leq \tau} F_x\left(R_i\right) - \sum_{i=1}^\tau c_i
  \right]
  = \E_\pi\left[
    \max_{1 \leq i \leq \tau} F_x\left(R_i\right) - \lambda_{\text{gen}}\sum_{i=1}^\tau N_i - \lambda_{\text{lat}} \sum_{i=1}^\tau T_i
  \right]
\end{equation}
where $\tau \in \{1, 2\}$ is the number of batches, and $\E_\pi$ is the expectation under policy $\pi$.

\subsection{The latency-aware policy} \label{sec:latency-aware-policy}
Here we describe the optimal policy that maximizes \eqref{eq:value}, which we call the \emph{latency-aware} policy.
\paragraph{Second-stage decision.} We will start with deriving the optimal decision of whether to generate the second batch, and if so at what size $N_2$. Denote $Q_1(s_1, a_1)$ as the Q-function of taking action $a_1$ in state $s_1$; clearly $Q_1(s_1, \mathrm{stop}) = F_x(R_1)$, and
\begin{equation*}
  Q_1(s_1, N_2) = - \lambda_{\text{gen}} N_2 - \lambda_{\text{lat}} \widehat T_2(s_1, N_2) + \E[\max\{F_x(R_1), F_x(R_2)\} \mid s_1, \, N_2]
\end{equation*}
where $\widehat T_2(s_1, N_2) \coloneqq \E[T_2 \mid s_1, N_2]$ is the expected latency of the second batch given everything observed after the first. In practice it may be estimated by regression from server logs, and we use its exact value in our simulation-based experiments (\cref{sec:experiments:setup}). Now, define $f_1(s_1, N_2) \coloneqq Q_1(s_1, N_2) - Q_1(s_1, \mathrm{stop})$. We can compute it as:
\begin{align*}
  f_1(s_1, N_2)
  &= - \lambda_{\text{gen}} N_2 - \lambda_{\text{lat}} \widehat T_2(s_1, N_2) + \E[\max\{F_x(R_1), F_x(R_2)\} \mid s_1, \, N_2] - F_x(R_1) \\
  &= - \lambda_{\text{gen}} N_2 - \lambda_{\text{lat}} \widehat T_2(s_1, N_2) + \E[\left[F_x(R_2) - F_x(R_1)\right]_+ \mid s_1, \, N_2]
\end{align*}
where $[\cdot]_+ = \max\{0,\cdot\}$.
The optimal policy for the second batch, which we call the \emph{latency-aware} policy, is as follows:
\begin{enumerate}
  \item Find $N_2^\star \in \arg\max_{N_2 \in \mathcal{Z}_2} f_1(s_1, N_2)$
  \item If $f_1(s_1, N_2^\star) > 0$, then generate a second batch of size $N_2^\star$. Otherwise, stop.
\end{enumerate}
Due to the CDF transformation resulting in uniform reward distributions ($F_x(r) \sim \Unif[0,1]$), the expected reward improvement term $\E[\left[F_x(R_2) - F_x(R_1)\right]_+ \mid s_1, \, N_2]$ in $f_1(s_1, N_2)$ can be computed in closed form as follows:
\begin{align*}
  \E[\left[F_x(R_2) - F_x(R_1)\right]_+ \mid s_1, \, N_2]
  &= \E[\left[F_x(R_2) - F_x(R_1)\right]_+ \mid F_x(R_1), \, N_2] 
  = \EI(F_x(R_1), N_2)
\end{align*}
where $\EI(u,N)$ is defined for $u \leq 1$ as follows (the third term vanishes when $u<0$):
\begin{equation} \label{eq:EI}
  \EI(u, N) \coloneqq \UE_{U_1, \ldots, U_N \simiid \Unif[0,1]}
  \left[\left(\max_{i \in [N]} U_i - u\right)_+\right]
  = \frac{N}{N+1} - u + \frac{u^{N+1}}{N+1} \mathbf{1}_{u \geq 0}.
\end{equation}
In practice, the CDF $F_x$ is unknown, so $\EI(F_x(R_1), N_2)$ must be estimated (\cref{sec:cdf-computation}). We call the policies that use the true $\EI(F_x(R_1), N_2)$ the \emph{oracle} policies.

\paragraph{Selection of $N_1$.} For both the latency-aware and latency-unaware policies $\pi$ (\cref{sec:batch_size_variants}), the first batch size is pre-set to maximize the profit:
\begin{equation*}
   N_1 \in \arg\max_{N_1 \in \mathcal{Z}_1} Q^\pi_0(s_0, N_1)
\end{equation*}
where $Q^\pi_0$ is the profit-to-go of policy $\pi$ from the empty initial state $s_0$ with initial batch size $N_1$. By the Bellman equation,
\begin{equation*}
  Q^\pi_0(s_0, N_1) = - \lambda_{\text{gen}} N_1 - \lambda_{\text{lat}} \overline T_1(N_1) + \E\left[Q_1(s_1, \pi_1(s_1)) \mid s_0, \, N_1 \right],
\end{equation*}
where $\overline T_1(N_1) \coloneqq \E[T_1 \mid s_0, N_1]$ is the expected latency of a first batch of size $N_1$, estimable in practice from historical latency logs. We estimate $Q^\pi_0(s_0, N_1)$ using numerical integration and Monte Carlo sampling, assuming the oracle policy is followed; details are given in \cref{apx:optimal-N1}.

\subsection{Baselines: latency-unaware and fixed batch size policies}
\label{sec:batch_size_variants}

For baseline comparison, we compare to a \emph{latency-unaware} policy, which does not have access to the latency measurement $T_1$ and therefore proceeds the same as the latency-aware policy but replaces $\widehat T_2(s_1, N_2)$ with $\overline T_2(N_2) \coloneqq \E[T_2 \mid N_2]$ in $f_1(s_1, N_2)$, using $f_1^{\text{lu}}(s_1, N_2)$ instead:
\begin{equation*}
  f_1^{\text{lu}}(s_1, N_2) = - \lambda_{\text{gen}} N_2 - \lambda_{\text{lat}} \overline T_2(N_2) + \EI(F_x(R_1), N_2).
\end{equation*}
Like existing adaptive Best-of-$N$ methods, the latency-unaware policy stops once its signal, here $F_x(R_1)$, exceeds a fixed threshold; in contrast, the latency-aware policy's threshold decreases with $T_1$ (\cref{apx:threshold,fig:overview_rule}).

We propose an \emph{adaptive batch size} policy, which uses $\mathcal{Z}_1 = \mathcal{Z}_2 = \mathcal{Z}$ for some chosen finite set $\mathcal{Z} \subseteq \mathbb{N}$ of allowed batch sizes; this allows the second batch size to be adaptively chosen. We compare it to its \emph{fixed batch size} baseline counterpart, meant to resemble the fixed-threshold rules of prior works \citet{fu2025dynasor,manvi2024adaptive,wan2025beacon}, which pre-commit to the second batch size: it sets $\mathcal{Z}_1 = \{N_1\}$ and $\mathcal{Z}_2 = \{N_2\}$, where $N_1$ and $N_2$ are jointly selected from the same set $\mathcal{Z}$ to maximize the profit in advance:
\begin{equation*}
  (N_1, N_2) \in \arg\max_{N_1, N_2 \in \mathcal{Z}}\; Q^\pi_0(s_0, N_1)\big|_{\mathcal{Z}_2 = \{N_2\}}.
\end{equation*}

\subsection{Estimating EI with shifted exponential and predicted Beta}
\label{sec:cdf-computation}
We estimate $F_x$ from the first batch's rewards by an estimated CDF $\hat F_x$, fitting a \emph{shifted-exponential} model to the exponentiated first-batch rewards above their empirical $\alpha$-quantile, for a tail level $\alpha$ (we use $\alpha = 1/2$), following the approach used by \citet{kalayci2025optimal}. Because this model is fitted to $e^r$, it is not invariant to the scale of the reward, and reward models differ in scale by orders of magnitude; we therefore first rescale rewards to unit median within-prompt standard deviation, using one constant estimated offline. \Cref{apx:cdf_estimation} defines the estimator and compares it with five alternatives, and \cref{apx:reward_scale} treats the rescaling.

A plug-in approach uses $\EI(\hat F_x(R_1), N_2)$. This falls short in two ways. First, $\hat F_x(R_1)$ overestimates $F_x(R_1)$ (\cref{apx:cdf_estimation}), so the plug-in underestimates $\EI(F_x(R_1), N_2)$ and the policy draws a second batch less often than it should. Second, removing this bias of $\hat F_x(R_1)$ is not enough: $\EI(u, N)$ is convex in $u$, so by Jensen's inequality
\begin{equation*}
  \EI\big(\E[F_x(R_1) \mid \hat F_x(R_1), N_1], N_2\big) \leq \E\big[\EI(F_x(R_1), N_2) \mid \hat F_x(R_1), N_1\big],
\end{equation*}
and the debiased plug-in is still too low.

We therefore predict $\E[\EI(F_x(R_1), N_2) \mid \hat F_x(R_1), N_1]$ directly, by predicting a Beta distribution $\mathrm{Beta}(a, b)$ for $F_x(R_1) \mid \hat F_x(R_1), N_1$. Since $\EI(u, N)$ is a polynomial in $u$ and $\UE_{U \sim \mathrm{Beta}(a, b)}[U^k] = \prod_{j=0}^{k-1} \frac{a+j}{a+b+j}$, its expectation follows in closed form from the Beta's moments:
\begin{equation*}
  \UE_{U \sim \mathrm{Beta}(a, b)}\left[\EI(U, N)\right] = \frac{N}{N+1} - \frac{a}{a+b} + \frac{1}{N+1} \prod_{j=0}^{N} \frac{a+j}{a+b+j},
\end{equation*}
which the policy uses in place of $\EI(F_x(R_1), N_2)$, for every $N_2$ at once. We call this estimate the \emph{predicted Beta}. Its parameters $a$ and $b$ are matched to a mean and a variance of $F_x(R_1)$ predicted by gradient-boosted trees, which take as inputs, besides $N_1$ and $\hat F_x(R_1)$, a few other CDF estimates $\hat F_{x,\alpha_1}(R_1), \ldots, \hat F_{x,\alpha_k}(R_1)$ (the shifted-exponential estimator at other tail levels $\alpha_1, \ldots, \alpha_k$) and the first batch's reward quantiles; the conditioning above then includes these inputs too. The trees are fitted offline on first batches drawn from other prompts with many scored responses each, with the normalized rank of $R_1$ among a prompt's responses (\cref{apx:quality_score}) standing in for $F_x(R_1)$. Online, the prediction takes negligible time compared to generating a batch. Most of the improvement over the plug-in comes from debiasing and from the additional inputs; predicting the whole Beta distribution rather than only its mean adds a smaller one. \Cref{apx:calibration} gives the details and compares the predicted Beta with the plug-in and with estimates that only debias or only add inputs.

\section{Experiments}
\label{sec:experiments}
We evaluate our method by simulation. \Cref{sec:experiments:setup} describes the experimental setup, including the latency model and definition of generations-equivalent cost; \cref{sec:gec_gamma} varies the cost of latency at one latency model setting, comparing our policy with the latency-unaware policy, with fixed batch sizes, and with Best-of-$N$; and \cref{sec:latency_model_results} varies the latency model.

\subsection{Experimental setup}
\label{sec:experiments:setup}
Every experiment below is run on 24 \emph{configurations}, each a combination of a language model, a reward model and a prompt dataset: four instruction-tuned language models, Llama-3.1-8B-Instruct, Gemma-2-9B-it, Mistral-7B-Instruct-v0.3 and Qwen2.5-7B-Instruct (Llama, Gemma, Mistral and Qwen for short); three reward models, FsfairX-LLaMA3-RM-v0.1, RM-Mistral-7B and ArmoRM-Llama3-8B-v0.1 (FsfairX, RM-Mistral and ArmoRM); and 100 prompts from each of two prompt datasets, AlpacaFarm \citep{dubois2023alpacafarm} and HH-RLHF \citep{hh-rlhf}. \Cref{apx:assets} gives the sources and licenses of the models and datasets. Every figure and table in this section reports the \emph{average} over the 24 configurations, so that no result rests on a single reward-distribution shape. \Cref{apx:scope} reports each configuration individually; \cref{apx:lambda-selection} discusses why one $(\lambda_{\text{gen}}, \lambda_{\text{lat}})$ pair is meaningful across all of them. In both datasets, each prompt $x_b$, $b=1,\ldots,100$, has a precomputed pool of 960 rewards $\mathcal{R}_b = \{r_{b,1}, \ldots, r_{b,960}\}$ sampled from $D_{r,\llm(x_b)}$. For evaluation, since the true CDFs are unknown, we approximate $F_{x_b}(r_{b,i})$ by the \emph{normalized rank} of $r_{b,i}$ within $\mathcal{R}_b$, defined as $\frac{k}{961}$ where $k$ is the rank of $r_{b,i}$ in $\mathcal{R}_b$, an unbiased estimate of $F_{x_b}(r_{b,i})$ (\cref{apx:quality_score}).

All policies choose their batch sizes from $\mathcal{Z} = \{1, \dots, 320\}$; the choice of 320 allows two batches to fit in a prompt's pool of 960 and covers typical batch sizes considered in batched adaptive Best-of-$N$ works \citep{raman2025abon,sun2024fast}.
For each $(\lambda_{\text{gen}}, \lambda_{\text{lat}})$ pair and policy, we run $n_{\text{MC}}=1000$ Monte Carlo trials per prompt $x_b$. Trial $i$ draws latencies $T_{1,i}$ and $T_{2,i}$ using the latency model described below, and a Stage~1 set $A_{b,i}$ of $N_1$ rewards without replacement from $\mathcal{R}_b$; if the policy continues with $N_{2,i}$ more, the second batch $B_{b,i}$ is drawn without replacement from $\mathcal{R}_b \setminus A_{b,i}$.

\paragraph{Latency model.} Given the batch sizes $N_1$ and $N_2$, batch $i$ takes $T_i = h(N_i)\,L_i$, where the \emph{base latencies} $L_1, L_2$ are drawn independently of the rewards from
\begin{equation}
\begin{pmatrix} \log L_1 \\ \log L_2 \end{pmatrix}
\sim \mathcal N\!\left(
\begin{pmatrix} \mu \\ \mu \end{pmatrix},\;
\sigma^2 \begin{pmatrix} 1 & \rho \\ \rho & 1 \end{pmatrix}
\right),
\qquad h(N) = 1 - \beta + \beta \frac{N}{80}.
\label{eq:latency_model}
\end{equation}
We fix the mean at $\overline L = \E[L_i] = 5$\,s and vary the coefficient of variation $c_{\mathrm{v}} = \mathrm{std}(L_i)/\overline L = \mathrm{std}(T_i)/\E[T_i]$; these relate to the log-normal parameters by $\mu = \log \overline L - \sigma^2/2$ and $\sigma = \sqrt{\log(1+c_{\mathrm{v}}^2)}$.
The factor $h(N)$ splits a batch's latency into a share $1 - \beta$ that does not depend on its size and a share $\beta$ proportional to it: $\beta = 0$ is latency independent of batch size, as when a batch's responses are decoded in parallel, and $\beta = 1$ latency proportional to it, as when a full server decodes them in turn (\cref{sec:serving}).
$\rho$ is the correlation of $\log T_1$ and $\log T_2$.
The latency-aware policy predicts $\widehat T_2(s_1, N_2) = h(N_2) \exp\bigl(\mu + \rho\,(\log L_1 - \mu) + (1 - \rho^2)\sigma^2/2\bigr)$ with $L_1 = T_1/h(N_1)$, which is $T_1 h(N_2)/h(N_1)$ at $\rho = 1$, and the latency-unaware policy predicts $\overline T_2(N_2) = \overline L h(N_2)$.

\paragraph{Policies.} \emph{Our policy} is the adaptive batch size, latency-aware policy, estimating $\EI$ using the predicted Beta (\cref{sec:cdf-computation}), where in each configuration, the regressors are fitted on half of its prompts and applied to the other half, and vice versa. We compare it with the latency-unaware policy, with the fixed batch size variants of both (\cref{sec:batch_size_variants}), and with Best-of-$N$ (which has expected latency $\overline L h(N)$).
Each configuration's Best-of-$N$ reference is computed on its own reward pool.
When $\lambda_{\text{gen}}$ is small, the $N$ that maximizes Best-of-$N$'s $V$ can exceed $320$, and the comparison would then be decided by the batch limit rather than by the policies; we therefore take 31 geometrically spaced values of $\lambda_{\text{gen}} \in [10^{-5}, 10^{-2}]$, where this $N$ is at most $320$ at every $\lambda_{\text{lat}}$ and $\beta$: at $\beta = 0$, Best-of-$N$'s latency does not depend on $N$, so this $N$ depends on $\lambda_{\text{gen}}$ alone and is at most $318$, its value at $\lambda_{\text{gen}} = 10^{-5}$; at $\beta > 0$, the latency grows with $N$, so a positive $\lambda_{\text{lat}}$ only lowers this $N$. \Cref{apx:heatmaps} shows our policy's mean number of generations, mean latency and mean normalized reward rank over these values of $\lambda_{\text{gen}}$ and a grid of values of $\lambda_{\text{lat}}$.

We also compare our policies with their oracle counterparts, approximating it by replacing $F_{x_b}(R_1)$ in $\EI(F_{x_b}(R_1),N_2)$ by the normalized rank $\frac{k}{961}$. The gap between the policies using estimated EI and their oracle counterparts shows the realized cost of estimating $\EI(F_{x_b}(R_1), N_2)$.

\paragraph{Generations-equivalent cost.} The objective $V$ of \cref{sec:algorithm} is mean reward minus cost: $V = \overline{F_x(R)} - \lambda_{\text{gen}}\overline N - \lambda_{\text{lat}}\overline T$, where $\overline{F_x(R)}$, $\overline N$ and $\overline T$ are a policy's mean maximum CDF-transformed reward, total number of generations, and latency. With $\gamma = \lambda_{\text{lat}}/\lambda_{\text{gen}}$, this is $V = \overline{F_x(R)} - \lambda_{\text{gen}}(\overline N + \gamma\overline T)$, and we call $C = \overline N + \gamma\overline T$ the policy's \emph{generations-equivalent cost}.
$\gamma$ is in units of generations per second, so $\gamma\overline T$ is the latency expressed in generations. $C$ solves $\lambda_{\text{gen}}\overline N + \lambda_{\text{lat}}\overline T = \lambda_{\text{gen}} C$: it is the number of generations that, with no latency, would cost the same.
We fix $\gamma$, sweep $\lambda_{\text{gen}}$ along the ray $\lambda_{\text{lat}} = \gamma\lambda_{\text{gen}}$ with $\lambda_{\text{lat}} \leq 1$, and plot each policy's mean normalized reward rank against its generations-equivalent cost. Best-of-$N$'s curve is its mean normalized reward rank against $N + \gamma\overline L h(N)$, which increases in $N$. We compare a policy with Best-of-$N$ at Best-of-$N$'s reward: if Best-of-$N$ reaches a mean normalized reward rank $\overline{F_x(R)}$ at generations-equivalent cost $C_{\text{BoN}}$ and the policy reaches it at $C_{\text{policy}}$, the \emph{proportion of generations-equivalent cost saved} is $(C_{\text{BoN}} - C_{\text{policy}})/C_{\text{BoN}}$, reported in percent throughout.

We report means of normalized reward rank, latency and generations over all prompts and trials, as point estimates without error bars, which are hard to compute correctly for draws without replacement; all policies and $\lambda$ pairs share the same seeded draws, so comparisons are paired (\cref{apx:seeding}).

\subsection{Latency awareness and adaptive batch sizes at varying latency cost}
\label{sec:gec_gamma}

\begin{figure}[tbp]
  \centering
  \includegraphics[width=\textwidth]{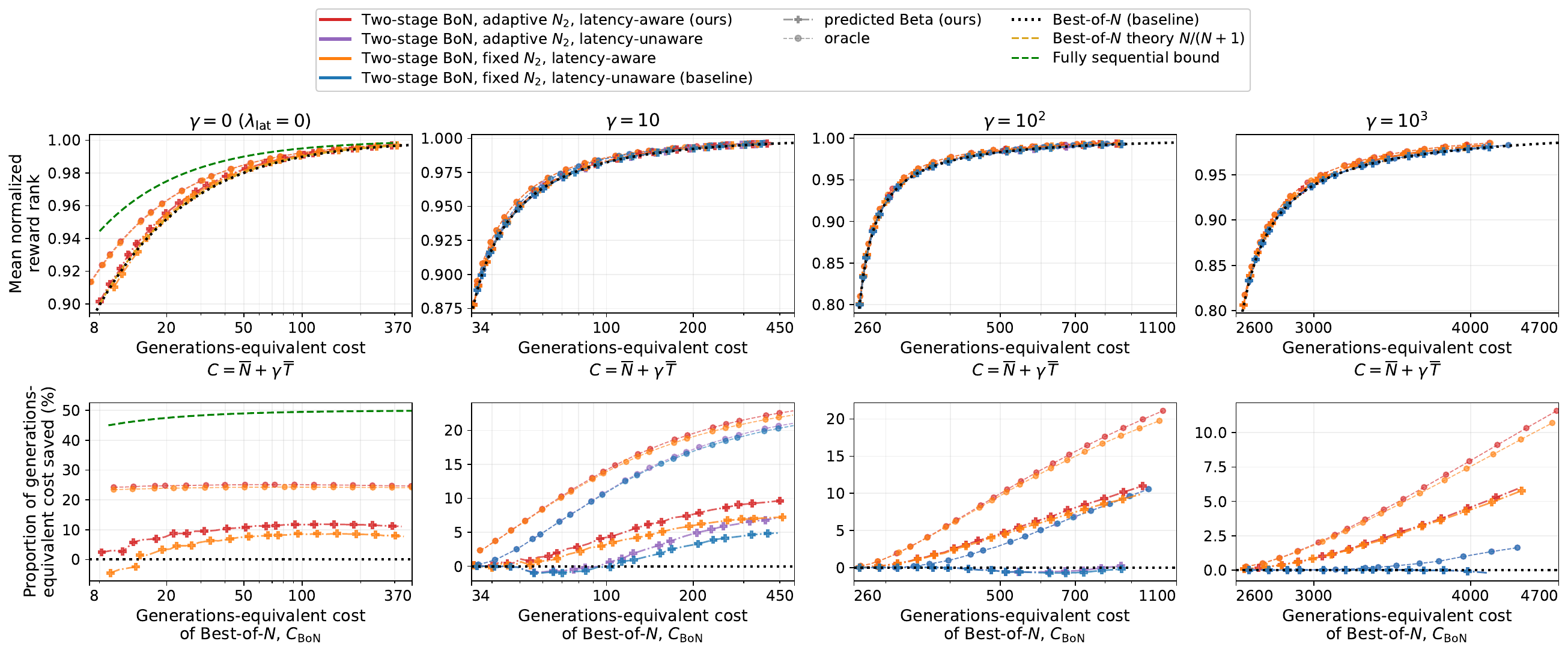}
  \caption{Top: mean normalized reward rank against generations-equivalent cost $C = \overline N + \gamma\overline T$ (mean generations $\overline N$, mean latency $\overline T$). Bottom: proportion of generations-equivalent cost saved $(C_{\text{BoN}} - C_{\text{policy}})/C_{\text{BoN}}$ at equal mean normalized reward rank against Best-of-$N$'s generations-equivalent cost $C_{\text{BoN}}$. $x$-axis in log scale. Lines are along a ray $\lambda_{\text{lat}} = \gamma\lambda_{\text{gen}}$ for $\gamma = 0$, $10$, $10^2$ and $10^3$ (columns), all at $\beta = 0.5$, $c_{\mathrm{v}} = 2$, $\rho = 1$, batch sizes $\{1, \dots, 320\}$, averaged over the 24 configurations. Colour is the policy (red: ours); dash-dot with plus signs: predicted Beta (ours); dashed with circles: oracle. Leftmost column shows $\gamma = 0$, where the latency-aware and latency-unaware policies coincide; Best-of-$N$'s theoretical reward $\overline N/(\overline N+1)$ in dashed gold, fully sequential bound (\cref{apx:fully_sequential}) in dashed green (mean reward $1 - \frac{1}{2 \overline N}$, proportion saved $\frac12 - \frac{1}{2 \overline N}$). Results for $\beta \in \{0,0.5,1\}$, $\gamma \in \{1, 10, 10^2, 10^3\}$ in \cref{apx:gec_grid}; per-configuration results in \cref{apx:scope}.}
  \label{fig:gec_gamma}
\end{figure}

\begin{table}[t]
\caption{Proportion (\%) of generations-equivalent cost saved at the reward of Best-of-$N$ with $N = 40$, by the adaptive batch size policies with batch sizes $\{1, \dots, 320\}$, at $c_{\mathrm{v}} = 2$ and $\rho = 1$, averaged over the 24 configurations. At $\gamma = 0$ the two latency rules are one policy and $\beta$ does not apply.}
\label{tab:gec_body}
\centering
\small
\begin{tabular}{lrrrrr}
\toprule
 &  & \multicolumn{2}{c}{predicted Beta (ours)} & \multicolumn{2}{c}{oracle} \\
\cmidrule(lr){3-4} \cmidrule(lr){5-6}
$\beta$ & $\gamma$ & latency-unaware & latency-aware & latency-unaware & latency-aware \\
\midrule
-- & $0$ & $10.1$ & $10.1$ & $25.0$ & $25.0$ \\
$0$ & $10^{1}$ & $-0.3$ & $0.5$ & $1.8$ & $4.1$ \\
$0$ & $10^{2}$ & $0.0$ & $0.0$ & $0.0$ & $0.1$ \\
$0.5$ & $10^{1}$ & $-0.5$ & $2.6$ & $7.8$ & $11.3$ \\
$0.5$ & $10^{2}$ & $-0.1$ & $2.6$ & $1.3$ & $6.5$ \\
$1$ & $10^{1}$ & $10.3$ & $14.0$ & $24.9$ & $28.2$ \\
$1$ & $10^{2}$ & $9.9$ & $28.2$ & $24.7$ & $39.6$ \\
\bottomrule
\end{tabular}%

\end{table}

\Cref{fig:gec_gamma} fixes the latency model at $\beta = 0.5$, $c_{\mathrm{v}} = 2$ and $\rho = 1$. Results for $\beta=0.5$ are shown here because we expect a typical server to exhibit a dependence of latency on batch size somewhere in between the extremes of $\beta = 0$ and $\beta = 1$.
At every reward level, our policy reaches Best-of-$N$'s reward with less generations-equivalent cost $C$, and saves at least as much as its latency-unaware counterpart.
Depending on $\gamma$ and $\beta$, the policies' savings generally either grow with $C$, or are the same across $C$.
As $\gamma$ grows, the latency-unaware policy's saving disappears: at $\gamma = 10^2$ and $10^3$, with the predicted Beta, it is close to Best-of-$N$ along the whole curve.
The adaptive batch size policies save slightly more than their fixed batch size counterparts, differing mainly when $\gamma \leq 10$, or when $\beta=1$ (see \cref{apx:gec_grid}).

At $\gamma = 0$ there is no latency cost and the generations-equivalent cost is the number of generations. As shown in \cref{tab:gec_body}, at $N = 40$ our policy uses $10.1\%$ less generations-equivalent cost than Best-of-$N$, compared to the oracle $25.0\%$ and the fully sequential bound of $48.75\%$ ($\frac{1}{2} - \frac{1}{2N}$ at $N = 40$). %

\subsection{How the benefit depends on the latency model}
\label{sec:latency_model_results}

\begin{figure}[t]
  \centering
  \includegraphics[width=\textwidth]{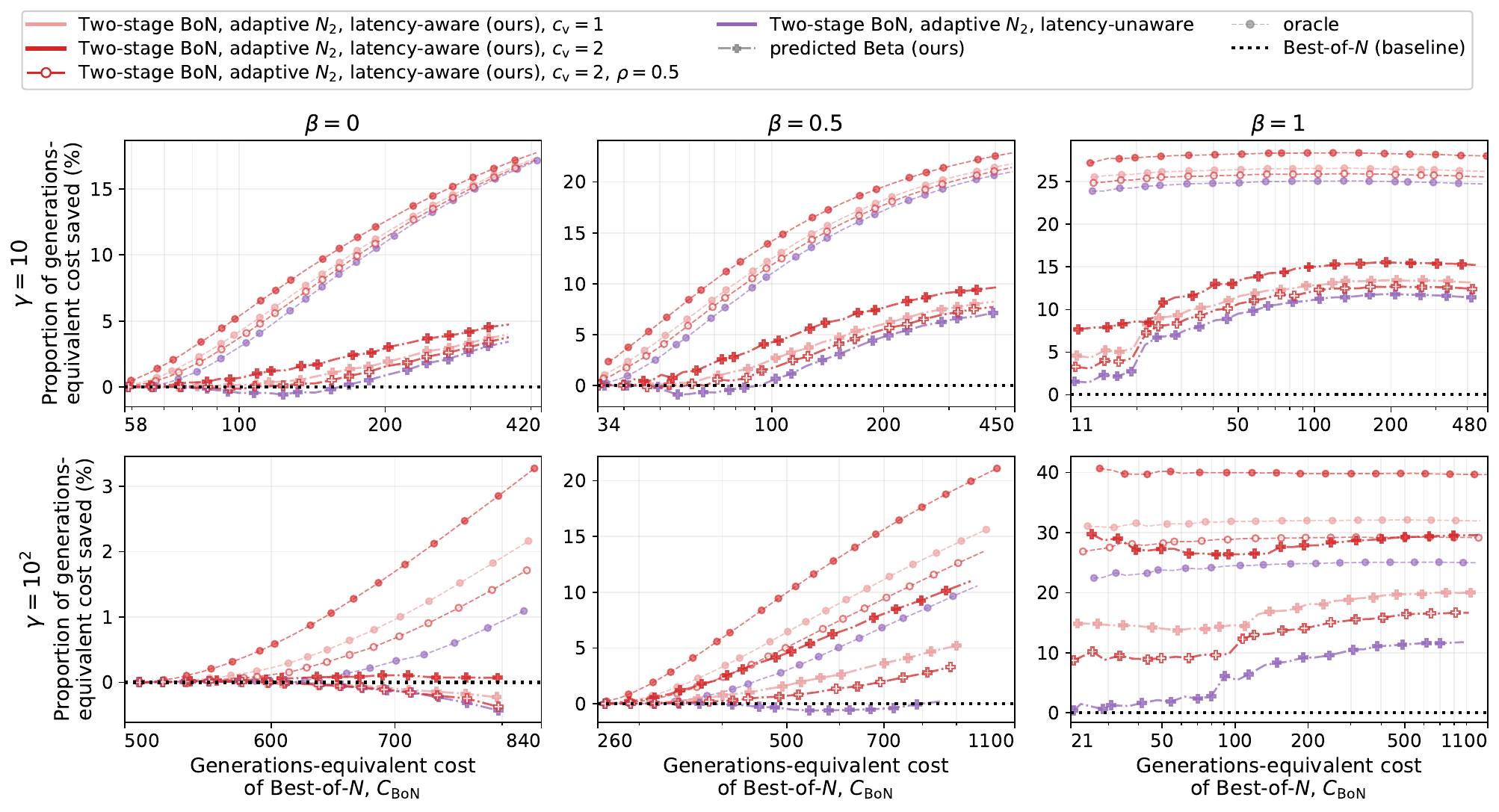}
  \caption{The proportion of generations-equivalent cost $C = \overline N + \gamma\overline T$ (mean generations $\overline N$, mean latency $\overline T$) saved, $(C_{\text{BoN}} - C_{\text{policy}})/C_{\text{BoN}}$ at equal mean normalized reward rank, against Best-of-$N$'s generations-equivalent cost $C_{\text{BoN}} = N + \gamma\overline L h(N)$ (log scale), at $\gamma = 10$ and $10^2$ (rows) and $\beta = 0$, $0.5$ and $1$ (columns), adaptive batch sizes $\{1, \dots, 320\}$, averaged over the 24 configurations. Our policy (red) is shown at $c_{\mathrm{v}} = 1$ and $2$ with $\rho = 1$, and at $c_{\mathrm{v}} = 2$ with $\rho = 0.5$. Since it never sees $T_1$, the latency-unaware policy (purple) behaves the same at every $c_{\mathrm{v}}$ and $\rho$ so only needs one line.}
  \label{fig:gec_beta}
\end{figure}

\Cref{fig:gec_beta} plots the adaptive batch size policies at various latency model settings and shows that the latency-aware policy's advantage over the latency-unaware policy is higher when latency grows more with batch size (high $\beta$), is highly variable (high $c_{\mathrm{v}}$), and is highly correlated between batches (high $\rho$).
\Cref{apx:corr_sweep} sweeps $\rho$ and compares with a misspecified latency predictor that assumes $\rho=1$.

\Cref{tab:gec_body} shows the same results at Best-of-$40$'s reward.
For both the latency-unaware and latency-aware policies, the saving generally increases with increasing $\beta$.
Furthermore, comparing absolute differences in proportional savings, the saving of the latency-aware policy over the latency-unaware policy generally increases with increasing $\beta$ for the same $\gamma$, and more so for higher $\gamma$.

\section{Conclusion}
\label{sec:conclusion}

We introduced a two-stage adaptive Best-of-$N$ algorithm that uses the first batch's observed rewards and its latency $T_1$, an informative predictor of the second batch's latency, to decide whether to draw a second batch and at what size. A CDF transformation of the rewards yields scale-free tradeoff parameters that require no per-task retuning, and a regression fitted offline on other prompts predicts a distribution for the CDF value of the first batch's best reward, from which the expected improvement follows in closed form. Empirically, our method reaches Best-of-$N$'s reward at lower generations-equivalent cost, and conditioning on $T_1$ saves at least as much of that cost as fixed-threshold (latency-unaware) baselines, and more when latency is highly variable, when it grows with batch size, and when the second batch's latency is highly predictable from the first's. Our experiments are simulation-based. An evaluation on a live serving stack with continuous batching \citep[vLLM;][]{kwon2023vllm} would show how well the latency assumptions of \cref{sec:serving} hold -- that one batch's latency predicts the next's, how much latency grows with batch size, and that adaptive requests like ours are too small a share of the load to shift what others face -- and identify the regimes in which our method helps. Fully sequential stopping, outcome-based reward models, and conditioning stopping decisions on measured latency in other test-time compute methods, such as chain-of-thought, are also future work.

\subsubsection*{AI use statement}
In this work, we used generative AI tools (large language model coding and
writing assistants) extensively. For software, most of the simulation code,
analysis scripts, and figure and table scripts were written by AI under our
direction; we did not review all of it line by line. For writing, AI drafted
and revised most of the manuscript text, including the related work and
appendices, and we edited it. AI also assisted in locating related work and
preparing bibliography entries, which we checked, with AI assistance, for
existence and correct authors, title, and venue.

The research question was developed with AI assistance; the idea of latency
adaptivity, the algorithm, and the main-text experiments are the authors' own.
The derivations in the appendix were mostly ours, but some of the proofs were
conceived by AI, and some of the appendix studies were designed with or by
AI. We did not use generative AI to generate or alter experimental data. We
take responsibility for the final content of this work.

\subsubsection*{Ethics statement}

This work is a simulation study that reuses publicly released prompts, model
generations, and reward-model scores from the AlpacaFarm and HH-RLHF datasets
(\cref{apx:assets}); it involves no human subjects, no new data collection, and
no release of new model weights or generations. The method allocates
inference-time compute and does not change what a language model is able to
produce, so it does not introduce new capability risks. Two considerations are
worth naming. First, the algorithm optimizes a reward model, and Best-of-$N$
selection inherits whatever biases that reward model carries; our contribution
is orthogonal to reward-model quality and does not mitigate such biases.
Second, by conditioning on server load, the method makes the amount of compute a
request receives depend on when it arrives, so response quality can vary across
users for reasons unrelated to their request; a deployment that cares about
equal treatment should account for this. The externality of one request's
continuation decision on other users' latency is assumed negligible when such requests are a small share of the server's
load (\cref{sec:serving}); testing this on a live serving stack is
future work (\cref{sec:conclusion}).

\subsubsection*{Reproducibility statement}

All of our results are simulations over precomputed reward pools, and are fully
reproducible on CPU. The algorithm and the closed-form quantities it uses are
specified in \cref{sec:algorithm,sec:cdf-computation}, with the derivation of
the Q-functions and the optimal policy in \cref{sec:latency-aware-policy}. The CDF
estimator and its baselines are specified in \cref{apx:cdf_estimation}. The
experimental setup -- datasets, language models, reward models, prompt and
pool sizes, the $(\lambda_{\text{gen}}, \lambda_{\text{lat}})$ grid, the number
of Monte Carlo trials, and the latency model -- is given in
\cref{sec:experiments:setup}, and the per-configuration results
behind the averages reported in the main text are in \cref{apx:scope}. The
datasets and the released reward scores we build on are documented in
\cref{apx:assets}; compute requirements are in \cref{apx:compute}.
\ifdefined\arxivpreprint\else
Our simulation code, including the scripts that regenerate every figure and table in
this paper from cached results, is submitted as supplementary material.
\fi

\bibliographystyle{iclr2027_conference}
\bibliography{common/bibliography}

\appendix

\crefalias{section}{appendix}
\crefalias{subsection}{subappendix}
\crefalias{subsubsection}{subsubappendix}

\AppendixFloatsBySection

\section{Optimal $N_1$ selection} \label{apx:optimal-N1}
The optimal first batch size is $N_1 \in \arg\max_{N_1 \in \mathcal{Z}_1} Q^\pi_0(s_0, N_1)$. For the latency-aware policy $\pi^\star$:
\begin{align*}
  Q^{\star}_0(s_0, N_1) = -\lambda_{\text{gen}} N_1 - \lambda_{\text{lat}} \overline T_1(N_1) + \E\left[V^\star_1(s_1) \mid s_0, \, N_{1} \right],
\end{align*}
where
\begin{align*}
  V^\star_1(s_1) &= \max\left\{F_x(R_1), \max_{N_2 \in \mathcal{Z}_2} Q_1(s_1, N_2)\right\} = F_x(R_1) + \left[\max_{N_2 \in \mathcal{Z}_2} f_1(s_1, N_2)\right]_+.
\end{align*}
For the latency-unaware policy $\pi_{\text{lu}}$:
\begin{align*}
  Q^{\pi_{\text{lu}}}_0(s_0, N_1) = - \lambda_{\text{gen}} N_1 - \lambda_{\text{lat}} \overline T_1(N_1) + \E\left[Q_1(s_1, \pi_{\text{lu}}(s_1)) \mid s_0, \, N_1 \right],
\end{align*}
where $\pi_{\text{lu}}(s_1) \in \arg\max_{a_1 \in \{\mathrm{stop}\} \cup \mathcal{Z}_2} Q_1^{\text{lu}}(s_1, a_1)$, defining $Q_1^{\text{lu}}(s_1, N_2) = F_x(R_1) + f_1^{\text{lu}}(s_1, N_2)$ for $N_2 \in \mathcal{Z}_2$ and $Q_1^{\text{lu}}(s_1, \mathrm{stop}) = F_x(R_1)$.

Here $Q^{\pi_{\text{lu}}}_0$ is written in terms of the true $Q_1$: the policy is evaluated at the second-batch latency it actually incurs, though its decision uses $\overline T_2(N_2)$. We show that $Q_1(s_1, \pi_{\text{lu}}(s_1))$ agrees in expectation with $Q_1^{\text{lu}}(s_1, \pi_{\text{lu}}(s_1))$ under the following condition on the latency model:
\begin{equation}
  \E\bigl[T_2 \bigm| s_0, \, N_1, \, N_2, \, r^{(1)}_1, \ldots, r^{(1)}_{N_1}\bigr] = \overline T_2(N_2),
  \qquad N_1 \in \mathcal{Z}_1, \; N_2 \in \mathcal{Z}_2.
  \label{eqn:lu-budget}
\end{equation}
Of the two independences in \eqref{eqn:lu-budget}, that from $N_1$ holds when requests from adaptive policies are a small share of the server's load (\cref{sec:serving}): the first batch completes before the second is submitted, so it can affect $T_2$ only through the state it leaves the server in, and one batch does not perturb that state. That from the rewards is a separate condition, response length being correlated with both reward and latency; it still leaves $T_2$ free to depend on $T_1$, which is the dependence the latency-aware policy uses. Both hold in our experiments, where, given $N_2$, the second batch's latency is drawn independently of the rewards and of $N_1$, and from the same distribution for every prompt.

Since $Q_1(s_1, \mathrm{stop}) = Q_1^{\text{lu}}(s_1, \mathrm{stop}) = F_x(R_1)$ and $Q_1(s_1, N_2) - Q_1^{\text{lu}}(s_1, N_2) = -\lambda_{\text{lat}}\bigl(\widehat T_2(s_1, N_2) - \overline T_2(N_2)\bigr)$ for $N_2 \in \mathcal{Z}_2$, the law of total expectation, splitting on the value of $\pi_{\text{lu}}(s_1)$, gives
\begin{align*}
  &\E\bigl[Q_1(s_1, \pi_{\text{lu}}(s_1)) - Q_1^{\text{lu}}(s_1, \pi_{\text{lu}}(s_1)) \bigm| s_0, N_1\bigr] \\
  &\quad = \Pr\bigl[\pi_{\text{lu}}(s_1) = \mathrm{stop} \bigm| s_0, N_1\bigr]\,
      \E\bigl[Q_1(s_1, \mathrm{stop}) - Q_1^{\text{lu}}(s_1, \mathrm{stop}) \bigm| \pi_{\text{lu}}(s_1) = \mathrm{stop}, \, s_0, N_1\bigr] \\
  &\quad\quad + \sum_{N_2 \in \mathcal{Z}_2} \Pr\bigl[\pi_{\text{lu}}(s_1) = N_2 \bigm| s_0, N_1\bigr]\,
      \E\bigl[Q_1(s_1, N_2) - Q_1^{\text{lu}}(s_1, N_2) \bigm| \pi_{\text{lu}}(s_1) = N_2, \, s_0, N_1\bigr] \\
  &\quad = - \lambda_{\text{lat}} \sum_{N_2 \in \mathcal{Z}_2} \Pr\bigl[\pi_{\text{lu}}(s_1) = N_2 \bigm| s_0, N_1\bigr]\,
      \E\bigl[\widehat T_2(s_1, N_2) - \overline T_2(N_2) \bigm| \pi_{\text{lu}}(s_1) = N_2, \, s_0, N_1\bigr] \\
  &\quad = - \lambda_{\text{lat}} \sum_{N_2 \in \mathcal{Z}_2} \Pr\bigl[\pi_{\text{lu}}(s_1) = N_2 \bigm| s_0, N_1\bigr]\,
      \Bigl( \E\bigl[\widehat T_2(s_1, N_2) \bigm| \pi_{\text{lu}}(s_1) = N_2, \, s_0, N_1\bigr] - \overline T_2(N_2) \Bigr) \\
  &\quad = 0.
\end{align*}
Conditioning on $\{\pi_{\text{lu}}(s_1) = N_2\}$ is not the same as conditioning on the second batch having size $N_2$: $f_1^{\text{lu}}$ does not involve $T_1$, so the event is a statement about the first-batch rewards, and conditioning on it reweights anything correlated with them. Condition \eqref{eqn:lu-budget} is what rules this out, holding as it does at every value of the rewards; averaging it over those values for which $\pi_{\text{lu}}(s_1) = N_2$ therefore returns the same value, which gives the last equality above:
\begin{align*}
  &\E\bigl[\widehat T_2(s_1, N_2) \bigm| \pi_{\text{lu}}(s_1) = N_2, \, s_0, N_1\bigr] \\
  &\quad = \E\bigl[\E[T_2 \mid s_0, r^{(1)}_{1:N_1}, N_1, T_1, N_2] \bigm| \pi_{\text{lu}}(s_1) = N_2, \, s_0, N_1\bigr] \\
  &\quad = \E\bigl[T_2 \bigm| \pi_{\text{lu}}(s_1) = N_2, \, s_0, N_1\bigr] \\
  &\quad = \E\Bigl[ \E\bigl[T_2 \bigm| s_0, N_1, N_2, r^{(1)}_1, \ldots, r^{(1)}_{N_1}\bigr]
    \Bigm| \pi_{\text{lu}}(s_1) = N_2, \, s_0, N_1 \Bigr] \\
  &\quad = \overline T_2(N_2).
\end{align*}
The first equality is the definition of $\widehat T_2(s_1, N_2)$; the second and third are the law of total expectation, the event $\{\pi_{\text{lu}}(s_1) = N_2\}$ being determined by the first-batch rewards; the fourth uses \eqref{eqn:lu-budget}.

Under \eqref{eqn:lu-budget},
\begin{equation*}
  Q^{\pi_{\text{lu}}}_0(s_0,N_1) = -\lambda_{\text{gen}} N_1 - \lambda_{\text{lat}} \overline T_1(N_1) + \E[V_1^{\text{lu}}(s_1) \mid s_0, N_1]
\end{equation*}
where $V_1^{\text{lu}}(s_1) \coloneqq F_x(R_1) + [\max_{N_2 \in \mathcal{Z}_2} f_1^{\text{lu}}(s_1, N_2)]_+$; the two policies' Q-functions ($Q^\star_0$ and $Q^{\pi_{\text{lu}}}_0$)
then have identical structure, differing only in whether the second-batch latency is $\widehat T_2(s_1, N_2)$ (latency-aware) or $\overline T_2(N_2)$ (latency-unaware). Otherwise $Q^{\pi_{\text{lu}}}_0$ must be evaluated as written, with the true $Q_1$ inside.

\paragraph{Computing $Q^\star_0(s_0, N_1)$ and $Q^{\pi_{\text{lu}}}_0(s_0, N_1)$ under the CDF transformation.}
These quantities are computed offline to determine the initial batch size $N_1$, and for the fixed batch size policies the second batch size $N_2$. Under the CDF transformation, and for the latency-unaware policy under \eqref{eqn:lu-budget}, both policies admit the decomposition
\begin{equation*}
  Q^\pi_0(s_0, N_1) \;=\; -\lambda_{\text{gen}} N_1 - \lambda_{\text{lat}} \overline T_1(N_1) + \frac{N_1}{N_1+1} + W^\pi_{N_1, \mathcal{Z}_2}, \qquad \pi \in \{\pi^\star, \pi_{\text{lu}}\},
\end{equation*}
where $\frac{N_1}{N_1+1} = \E[F_x(R_1) \mid s_0, N_1]$ is the expected maximum of $N_1$ i.i.d.\ $\Unif[0,1]$ rewards, and
\begin{align*}
  W^\star_{N_1, \mathcal{Z}_2} &\coloneqq \E\!\left[\left[\max_{N_2 \in \mathcal{Z}_2} f_1(s_1, N_2)\right]_+ \;\middle|\; s_0, N_1\right], &
  W^{\pi_{\text{lu}}}_{N_1, \mathcal{Z}_2} &\coloneqq \E\!\left[\left[\max_{N_2 \in \mathcal{Z}_2} f_1^{\text{lu}}(s_1, N_2)\right]_+ \;\middle|\; s_0, N_1\right].
\end{align*}
To compute these we specialize to the case simulated in \cref{sec:experiments}, in which the second batch's expected latency depends on $s_1$ only through $T_1$ and $N_1$, so that we may write it $\widehat T_2(T_1, N_1, N_2)$, and in which $T_1$ is independent of the first-batch rewards. Writing $p_{N_1}$ for the distribution of $T_1$ given $N_1$, and since $F_x(R_1)$ has density $N_1 u^{N_1-1}$ on $[0,1]$,
\begin{align*}
  W^\star_{N_1, \mathcal{Z}_2} &= \UE_{T_1 \sim p_{N_1}}\!\left[ \int_0^1 \left[\max_{N_2 \in \mathcal{Z}_2}\!\left\{ - \lambda_{\text{gen}} N_2 - \lambda_{\text{lat}} \widehat T_2(T_1, N_1, N_2) + \EI(u, N_2) \right\}\right]_+ N_1 u^{N_1 - 1} \, du \right], \\
  W^{\pi_{\text{lu}}}_{N_1, \mathcal{Z}_2} &= \int_0^1 \left[\max_{N_2 \in \mathcal{Z}_2}\!\left\{ - \lambda_{\text{gen}} N_2 - \lambda_{\text{lat}} \overline T_2(N_2) + \EI(u, N_2) \right\}\right]_+ N_1 u^{N_1 - 1} \, du,
\end{align*}
with the integrals over $u$ evaluated by quadrature and the expectation over $T_1$ by Monte Carlo over samples from $p_{N_1}$.

If $T_1$ is not independent of the first-batch rewards, or if $\widehat T_2$ depends on $s_1$ through more than $T_1$ and $N_1$, this factorization does not apply; one may then either model the distribution of $s_1$ given $N_1$ (the first-batch rewards together with $T_1$) and estimate $W^\star_{N_1, \mathcal{Z}_2}$ by Monte Carlo over draws of $s_1$, from which both $F_x(R_1)$ and $\widehat T_2(s_1, N_2)$ follow, or retain the factorization as an approximation. Dependence of this kind is plausible in deployment, since response length correlates with both reward and latency. Either way, only the offline choice of $N_1$ is affected: the online stopping rule is not affected.

\section{Connection to Pandora's Box and to fixed-threshold rules}
\label{apx:pandora}

\subsection{Fully sequential mean reward upper bound}
\label{apx:fully_sequential}
In the $\gamma = 0$ column of \cref{fig:gec_gamma}, the dashed green curve is the optimal \emph{fully sequential} (one-at-a-time) stopping policy on CDF-transformed rewards: in the top row its mean maximum reward $1 - \frac{1}{2N}$ against its expected number of generations $N$, and in the bottom row the proportion $\frac12 - \frac{1}{2N_{\text{BoN}}}$ of Best-of-$N$'s generations it saves at Best-of-$N$'s reward, derived below. This is the limiting case of the multi-stage extension of \cref{sec:algorithm} in which $\mathcal{Z}_j = \{1\}$ for every stage $j \geq 1$ and the horizon is unbounded -- i.e.\ the Pandora's Box problem with infinitely many i.i.d.\ boxes considered by \citet{kalayci2025optimal}. Since our two-stage Best-of-$N$ setting is a constrained version of this problem, no two-stage policy can exceed this curve.

For a fixed prompt $x$, write $U_i = F_x(r_i) \sim \Unif[0,1]$ for the CDF-transformed reward of generation $i$. With per-generation cost $c \in (0, 1/2)$, the optimal policy $\pi^\star$ is the Gittins index policy \citep{weitzman1979optimal,scully2025gittins}, which stops after $\tau_{\pi^\star, x}$ generations, the first time the best value drawn exceeds the Gittins index $G$ of a further generation, the solution of $\EI(G, 1) = c$, i.e.\ $G = 1 - \sqrt{2c}$. We can easily find that, in general, the optimal profit equals the Gittins index,
\begin{equation*}
  \profit{\pi^\star}[x] \;=\; \E\!\left[\max_{i \leq \tau_{\pi^\star, x}} U_i - c \, \tau_{\pi^\star, x}\right] \;=\; G \;=\; 1 - \sqrt{2c},
\end{equation*}
and $\tau_{\pi^\star, x} \sim \Geo(\sqrt{2c})$, so $\E[\tau_{\pi^\star, x}] = 1/\sqrt{2c}$. Rearranging,
\begin{equation*}
  \E\!\left[\max_{i \leq \tau_{\pi^\star, x}} U_i\right] \;=\; G + c \, \E[\tau_{\pi^\star, x}] \;=\; (1 - \sqrt{2c}) + \frac{c}{\sqrt{2c}} \;=\; 1 - \frac{\sqrt{2c}}{2}.
\end{equation*}
Let $\overline N_{\text{PB}}$ be the fully sequential policy's expected number of generations. Since it generates one generation at a time, we have $\overline N_{\text{PB}} = \E[\tau_{\pi^\star, x}]$. Substituting $\sqrt{2c} = 1/\overline N_{\text{PB}}$, this gives:
\begin{equation*}
  \E\!\left[\max_{i \leq \tau_{\pi^\star, x}} U_i\right] \;=\; 1 - \frac{1}{2 \overline N_{\text{PB}}}.
\end{equation*}
In the proportion-saved view, let $N_{\text{BoN}}$ be the $N$ used for Best-of-$N$. The fully sequential policy reaches Best-of-$N$'s mean reward $N_{\text{BoN}}/(N_{\text{BoN}}+1)$ when $1 - \frac{1}{2\overline N_{\text{PB}}} = \frac{N_{\text{BoN}}}{N_{\text{BoN}}+1}$, which gives
\begin{equation*}
  1 - \frac{\overline N_{\text{PB}}}{N_{\text{BoN}}} \;=\; \frac{1}{2} - \frac{1}{2N_{\text{BoN}}}.
\end{equation*}
This compares with Best-of-$N$'s theoretical reward $N/(N+1)$; \cref{fig:gec_gamma} measures the policies against the empirical Best-of-$N$, whose reward is slightly lower because of tied rewards (\cref{apx:quality_score}).

We include this curve in the plots as a \emph{reward-only} reference: it represents the most reward-adaptive policy possible -- evaluating generations one at a time and deciding after each whether to continue -- and so upper-bounds the mean reward attainable by any policy with the same expected number of generations $N$, including any two-stage policy considered in this paper. This reference says nothing about latency: a fully sequential policy that draws $N$ generations one at a time incurs roughly $N$ times the latency of a single batch of size $N$, which is precisely the regime our two-stage formulation is designed to avoid. The curve is therefore informative as a ceiling on the reward axis only, and the latency cost of actually attaining it is not reflected in the plot.

\subsection{The stopping rule is a threshold on $F_x(R_1)$}
\label{apx:threshold}
Write $c_{N_2} \coloneqq \lambda_{\text{gen}} N_2 + \lambda_{\text{lat}} \widehat T_2(s_1, N_2)$ for the latency-aware policy and $c_{N_2} \coloneqq \lambda_{\text{gen}} N_2 + \lambda_{\text{lat}} \overline T_2(N_2)$ for the latency-unaware policy, so that the continuation value of either is $\EI(F_x(R_1), N_2) - c_{N_2}$ (\cref{sec:algorithm}). The function $\EI(\cdot, N_2)$ (defined in \eqref{eq:EI}, \cref{sec:latency-aware-policy}) is continuous on $(-\infty,1]$ and, since its derivative is $-\Pr[\max_{i \in [N_2]} U_i > u] < 0$ for $u < 1$, with $U_1, \ldots, U_{N_2}$ as in its definition, strictly decreasing on $(-\infty, 1]$, from $+\infty$ to $\EI(1, N_2) = 0$. So, as $c_{N_2} > 0$, there is a unique $G_{N_2} < 1$ with $\EI(G_{N_2}, N_2) = c_{N_2}$: the Gittins index \citep{weitzman1979optimal,scully2025gittins} of a box in the Pandora's Box problem with opening cost $c_{N_2}$ whose value is the maximum of $N_2$ independent $\Unif[0,1]$ draws, the CDF values of the second batch's rewards.

\begin{proposition}
\label{prop:threshold}
Either policy stops after the first batch if and only if $F_x(R_1) \geq \max_{N_2 \in \mathcal{Z}_2} G_{N_2}$.
\end{proposition}
\begin{proof}
\begin{align*}
  \text{stop}
  &\iff \max_{N_2 \in \mathcal{Z}_2} \bigl[\EI(F_x(R_1), N_2) - c_{N_2}\bigr] \leq 0
  \iff \forall N_2 \in \mathcal{Z}_2,\; \EI(F_x(R_1), N_2) \leq c_{N_2} \\
  &\iff \forall N_2 \in \mathcal{Z}_2,\; F_x(R_1) \geq G_{N_2}
  \iff F_x(R_1) \geq \max_{N_2 \in \mathcal{Z}_2} G_{N_2}.
\end{align*}
The third equivalence holds because $\EI(\cdot, N_2)$ is strictly decreasing on $(-\infty, 1]$ and equals $c_{N_2}$ at $G_{N_2}$.
\end{proof}

For the latency-unaware policy, $c_{N_2}$ depends only on $\lambda_{\text{gen}}$, $\lambda_{\text{lat}}$ and $\overline T_2(N_2)$, so the threshold $\max_{N_2 \in \mathcal{Z}_2} G_{N_2}$ is fixed before any request arrives: this is the fixed-threshold stopping rule of existing adaptive Best-of-$N$ methods (\cref{sec:related}), applied to the CDF value of the best first-batch reward. Every threshold in $(0,1)$ is attained by some $(\lambda_{\text{gen}}, \lambda_{\text{lat}})$, so, up to the range and resolution of the grid, sweeping the tradeoff parameters as in \cref{sec:experiments:setup} traces the latency-unaware policy over all fixed thresholds on this signal. For the latency-aware policy, $c_{N_2}$, and with it $G_{N_2}$, depends on $s_1$ through $\widehat T_2(s_1, N_2)$, so its threshold moves from request to request; with the predictor of \cref{sec:experiments:setup}, which increases with $T_1$ when $\rho > 0$, the threshold falls as $T_1$ grows.

\section{Measuring response quality: the normalized reward rank}
\label{apx:quality_score}
We score a returned response by its \emph{normalized reward rank} within its prompt's pool. Let $k$ be the rank of $r_{b,i}$ among the $M = 960$ rewards of $\mathcal{R}_b$, with tied rewards given their average rank; the normalized reward rank is $k/(M+1)$. The oracle policies use this quantity in place of $F_{x_b}$.

Without ties, the normalized reward rank is an unbiased estimate of the true CDF value $F_{x_b}(r_{b,i})$. The pool is an i.i.d.\ sample from $F_{x_b}$, so $F_{x_b}$ evaluated at its $k$-th smallest reward is distributed as the $k$-th order statistic of $M$ i.i.d.\ $\Unif[0,1]$ variables, whose mean is $k/(M+1)$. For Best-of-$N$ and the oracle policies, which select by rank alone, the mean normalized reward rank is therefore an unbiased estimate of $\E_\pi[\max_{1 \leq i \leq \tau} F_x(R_i)]$, the reward term of the objective $V$; for the policies using estimated EI, which also use reward values, it is unbiased to close approximation.

The empirical Best-of-$N$ curve in the figures and tables is computed exactly rather than by simulation. Sort a prompt's normalized reward ranks as $u_{(1)} \leq \dots \leq u_{(M)}$. Best-of-$N$ draws $N$ of the $M$ responses without replacement and returns the best, and the best is the response of rank $K$ in this order; $K = k$ exactly when the draw contains response $k$ and $N-1$ of the $k-1$ responses below it. Its mean normalized reward rank is therefore
\begin{equation*}
  \E[u_{(K)}] = \sum_{k=N}^{M} u_{(k)} \, \binom{k-1}{N-1} \Big/ \binom{M}{N},
\end{equation*}
averaged over the prompts; tied rewards are simply equal entries of the sorted list.

The Best-of-$N$ theory curve $N/(N+1)$ follows directly. Without ties $u_{(k)} = k/(M+1)$, so the sum is $\E[K]/(M+1)$, and $\E[K] = N(M+1)/(N+1)$: the mean normalized reward rank is exactly $N/(N+1)$, for any pool size $M$. The rewards, however, are stored at bfloat16 precision, so a prompt's pool holds only $6$--$40\%$ distinct values on average, depending on the configuration. Average ranks are the mean of the scores over all tie-breaks, each of which gives $N/(N+1)$; since the maximum is convex in the scores, ties can only lower Best-of-$N$'s expected normalized reward rank. The effect is small: for $N \geq 80$ the empirical Best-of-$N$ curve lies below $N/(N+1)$ by about $0.001$ at most in any configuration, and by at most $0.0003$ in the average over the 24 configurations plotted in the figures.

\section{Estimating $F_x(R_1)$ prior to regression}
Our method estimates the CDF value $F_x(R_1)$ of the first batch's best reward $R_1 = \max_i r^{(1)}_i$ and uses the estimate as an input to a regressor that predicts the expected improvement in reward (\cref{sec:cdf-computation,apx:calibration}).
\Cref{apx:cdf_estimation} compares estimators of $F_x(R_1)$ for that role. \Cref{apx:reward_scale} describes our reward rescaling method for the shifted exponential estimator, a minor detail.

\subsection{Comparing $F_x(R_1)$ estimation methods}
\label{apx:cdf_estimation}
Here we compare estimators of $F_x(R_1)$ used as input to the regressor. For an informative comparison of their usefulness as regression inputs, we note that all of these estimators are biased, so we compare them both as they are and after regression-based debiasing (as described in \cref{apx:calibration}). The regressor learns the mean of $F_x(R_1)$ given its inputs, which removes their bias, so what matters is the accuracy in estimating $F_x(R_1)$ once its bias is removed: the RMSE after debiasing is a proxy for how useful an estimator is as an input for predicting the expected improvement, whose accuracy \cref{apx:calibration} measures directly.
We evaluate six candidate estimators of $F_x(R_1)$. Each fits a CDF $\hat F_x$ to the $N_1$ Stage-1 rewards and evaluates it at $R_1$:
\begin{itemize}
  \item \emph{Shifted exponential} (ours). The estimator our method uses (\cref{sec:cdf-computation}), fitted to the upper tail of the
    exponentiated Stage-1 rewards following \citet{kalayci2025optimal}. Concretely, fix a tail level $\alpha \in (0, 1)$ (we use $\alpha = 1/2$), let $s > 0$ be the reward scale of \cref{apx:reward_scale}, let $\hat q_\alpha$ denote the empirical $\alpha$-quantile of the rescaled rewards $r^{(1)}_1/s, \ldots, r^{(1)}_{N_1}/s$, and let $\theta = e^{\hat q_\alpha}$. Among the rewards above $\hat q_\alpha$, we fit to $e^{r^{(1)}_i/s}$ an exponential distribution shifted to start at $\theta$, with mean parameter $\hat\mu$ fitted by maximum likelihood:
    \begin{equation*}
      \hat\mu \;=\; \frac{1}{|\mathcal{I}|} \sum_{i \in \mathcal{I}} \left(e^{r^{(1)}_i/s} - \theta\right),
      \qquad
      \mathcal{I} = \{i : r^{(1)}_i/s > \hat q_\alpha\},
    \end{equation*}
    and define the estimator
    \begin{equation}
      \hat F_x(R_1) \;=\; \alpha + (1 - \alpha) \left(1 - \exp\!\left(-\frac{e^{R_1/s} - \theta}{\hat\mu}\right)\right).
      \label{eq:shifted_exp}
    \end{equation}
  \item \emph{In-sample KDE.} Gaussian kernel density estimator
    $\hat F_x(R_1) = \frac{1}{N_1} \sum_{i=1}^{N_1} \Phi\!\bigl((R_1 - r^{(1)}_i)/h\bigr)$
    with bandwidth $h = \hat\sigma\, N_1^{-1/5}$ (Scott's rule), where $\hat\sigma$ is the sample standard deviation of the $N_1$ Stage-1 rewards, evaluated \emph{using the same sample} that defined the bandwidth and the kernel centers. Following \citet{raman2025abon}, we treat the in-sample KDE variant as the canonical KDE baseline.
  \item \emph{Leave-one-out (LOO) KDE, $\times 0.5$ and $\times 1.0$.} Same as in-sample KDE, but the term corresponding to $R_1$ itself (which contributes a fixed $\Phi(0) = 1/2$) is dropped from the sum and the denominator becomes $N_1 - 1$. The two variants differ in the bandwidth multiplier ($\texttt{bw\_mult}=0.5$ and $\texttt{bw\_mult}=1.0$), and we set $h = \texttt{bw\_mult} \, \hat\sigma\, N_1^{-1/5}$.
  \item \emph{Normal.} $\hat F_x(R_1) = \Phi\!\bigl((R_1 - \bar r)/\hat\sigma\bigr)$, where $\bar r$ is the sample mean of the Stage-1 rewards.
  \item \emph{GMM-2.} A two-component Gaussian mixture model (GMM) is fit to all $N_1$ Stage~1 rewards by maximum likelihood using the expectation--maximization (EM) algorithm. EM is initialised by splitting the sorted sample at its median: each half's mean and standard deviation start one component, with equal weights. It then runs a fixed $30$ iterations, alternating between computing each reward's posterior probability of belonging to each component (E-step) and re-estimating the weights $w_k$, means $\hat\mu_k$, and standard deviations $\hat\sigma_k$ from those probabilities (M-step). The fitted mixture CDF is evaluated at the maximum, $\hat F_x(R_1) = \sum_{k=1}^{2} w_k\, \Phi\bigl((R_1 - \hat\mu_k)/\hat\sigma_k\bigr)$. Unlike the Normal fit, the mixture can represent skewed or bimodal reward distributions.
\end{itemize}

\paragraph{Evaluation setup.}
We run this evaluation separately in each of the 24 configurations of \cref{sec:experiments:setup}, each with 100 prompts, and average the results over the 24. For each prompt $x_b$, $b = 1, \ldots, B$, we partition the 960 precomputed rewards into a Stage-1 \emph{training pool} of size 480 and a held-out pool of size 480. For each estimator and each value of $N_1 \in \{15, 30, 60, 120, 240\}$, we run $M = 200$ independent Monte Carlo trials. In trial $m$, we draw a Stage-1 set of size $N_1$ without replacement from the training pool of each prompt, compute the per-prompt Stage-1 maximum $R_{1,b}^{(m)}$, evaluate the estimator $\hat F_{x_b}(R_{1,b}^{(m)})$, and compare it to the held-out empirical CDF value
\begin{equation*}
  \tilde F_{x_b}(R_{1,b}^{(m)}) \;\coloneqq\; \frac{|\{r \in \mathcal{H}_{x_b} : r \le R_{1,b}^{(m)}\}|}{|\mathcal{H}_{x_b}|},
\end{equation*}
where $\mathcal{H}_{x_b}$ is the held-out reward pool for prompt $x_b$. Splitting the rewards into disjoint training and held-out pools ensures that the held-out empirical CDF is an independent ground-truth estimate; in particular, a Stage-1 set of size $N_1 = 480$ drawn from the 480-reward training pool would be deterministic, eliminating the possibility of MC variation, which is why we cap the $N_1$ value at 240.
Each estimator is also evaluated after debiasing: for each estimator and each $N_1$, the binned-means map of \cref{apx:calibration} (the mean of $\tilde F_{x_b}(R_1)$ in each of $25$ equal-count bins of the estimate, interpolated linearly) is fitted on half of the configuration's prompts and applied to the other half, and vice versa.

\begin{figure}[htbp]
  \centering
  \includegraphics[width=\textwidth]{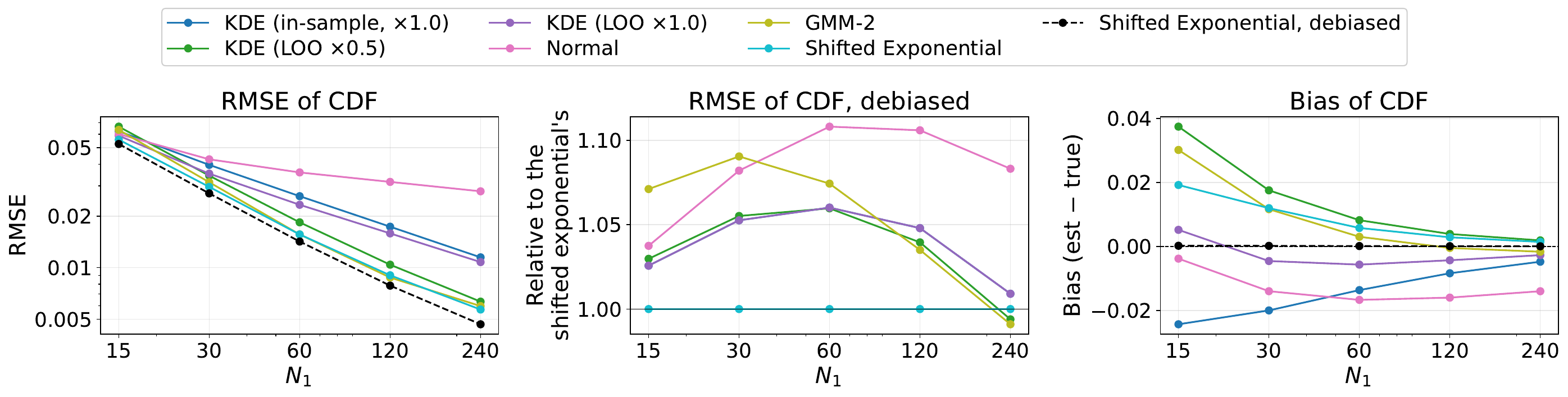}
  \caption{Six estimators of $F_x(R_1)$ against the Stage-1 batch size $N_1$ (log scale), averaged over the 24 configurations of \cref{sec:experiments:setup}: RMSE of the estimate (left), RMSE after debiasing divided by that of the debiased shifted exponential (middle), and bias of the estimate (right). Dashed black: the shifted exponential after debiasing. The in-sample KDE's debiased curve coincides with the leave-one-out KDE's ($\times 1.0$): with the same bandwidth, the former is an increasing affine function of the latter, which the debiasing map undoes. The in-sample KDE is the canonical KDE baseline of \citet{raman2025abon}; the predicted Beta takes the shifted-exponential estimate as its main input (\cref{apx:calibration}).}
  \label{fig:cdf_estimation}
\end{figure}

\Cref{fig:cdf_estimation} reports, against $N_1$, each estimator's RMSE (left), its RMSE after debiasing divided by that of the debiased shifted exponential (middle), and its bias (right), as point estimates without confidence bands, for the same reason as in \cref{sec:experiments:setup}; \cref{tab:cdf_estimator_rmse} reports the debiased RMSE and the bias. Both are averaged over the 24 configurations.

Every estimator is biased, overestimating or underestimating $F_x(R_1)$ depending on the estimator and on $N_1$ (\cref{tab:cdf_estimator_rmse}). Once debiased, the estimators differ less, and the shifted exponential has the lowest RMSE at every $N_1$ up to $120$ and at $N_1 = 240$ is within $0.01 \times 10^{-2}$ of the lowest. We therefore use the shifted exponential as the regressor's main input, also for its single closed form and its having no fitting procedure that can fail to converge.

\begin{table}[t]
\caption{RMSE of each estimator of $F_x(R_1)$ after debiasing, and bias of the estimate itself, against the held-out empirical CDF value, for the six estimators considered, as a function of $N_1$, averaged over the 24 configurations of \cref{sec:experiments:setup}. Held-out rewards tied with $R_1$ count as half. Positive bias: the estimator overstates $F_x(R_1)$. Averaged over 100 prompts, $M = 200$ Monte Carlo trials, and the 24 configurations; lowest debiased RMSE per column in bold, both entries where two tie at the printed precision. Each estimator is debiased by binned means fitted on the other half of the configuration's prompts. Per-configuration values are in Table~\ref{tab:scope_cdf_rmse}.}
\label{tab:cdf_estimator_rmse}
\centering
\small
\resizebox{\ifdim\width>\linewidth\linewidth\else\width\fi}{!}{%
\begin{tabular}{lrrrrr}
\toprule
RMSE of CDF, debiased ($\times 10^{-2}$) & $N_1 = 15$ & $30$ & $60$ & $120$ & $240$ \\
\midrule
In-sample KDE & 5.40 & 2.85 & 1.50 & 0.82 & 0.47 \\
LOO KDE ($\times 0.5$) & 5.42 & 2.86 & 1.50 & 0.82 & \textbf{0.46} \\
LOO KDE ($\times 1.0$) & 5.40 & 2.85 & 1.50 & 0.82 & 0.47 \\
Normal & 5.46 & 2.93 & 1.57 & 0.87 & 0.51 \\
GMM-2 & 5.64 & 2.96 & 1.52 & 0.81 & \textbf{0.46} \\
Shifted exponential (ours) & \textbf{5.27} & \textbf{2.71} & \textbf{1.42} & \textbf{0.78} & 0.47 \\
\midrule
Bias of CDF ($\times 10^{-2}$) & $N_1 = 15$ & $30$ & $60$ & $120$ & $240$ \\
\midrule
In-sample KDE & $-2.43$ & $-2.00$ & $-1.37$ & $-0.84$ & $-0.48$ \\
LOO KDE ($\times 0.5$) & $+3.74$ & $+1.75$ & $+0.82$ & $+0.39$ & $+0.19$ \\
LOO KDE ($\times 1.0$) & $+0.52$ & $-0.46$ & $-0.57$ & $-0.43$ & $-0.27$ \\
Normal & $-0.38$ & $-1.40$ & $-1.67$ & $-1.60$ & $-1.40$ \\
GMM-2 & $+3.01$ & $+1.16$ & $+0.30$ & $-0.05$ & $-0.17$ \\
Shifted exponential (ours) & $+1.92$ & $+1.19$ & $+0.58$ & $+0.28$ & $+0.14$ \\
\bottomrule
\end{tabular}
}%

\end{table}

\paragraph{Across language models, reward models, and datasets.}
\Cref{tab:scope_cdf_rmse} repeats the comparison of debiased RMSE in each of the 24 configurations of \cref{sec:experiments:setup}, as the ratio of the shifted exponential's to that of the best competing estimator. The shifted exponential has the lowest debiased RMSE in all 24 configurations at $N_1 = 15$ and $30$, in $22$ at $N_1 = 60$, in $16$ at $N_1 = 120$ and in $9$ at $N_1 = 240$, where the leave-one-out KDE with the narrower bandwidth and the Gaussian mixture are most often ahead.

\begin{table}[t]
\caption{RMSE after debiasing of the shifted-exponential estimate of $F_x(R_1)$ divided by that of the best of the five competing estimators after the same debiasing, in each of the 24 configurations of \cref{sec:experiments:setup}. Below one, ours is the most accurate; bold marks those entries. Averaged over 100 prompts and $M = 200$ Monte Carlo trials. Debiased, our estimator has the lowest RMSE in 24, 24, 22, 16, 9 of the 24 configurations at $N_1 = 15, 30, 60, 120, 240$ respectively.}
\label{tab:scope_cdf_rmse}
\centering
\small
\begin{tabular}{lllrrrrr}
\toprule
dataset & RM & LM & $N_1 = 15$ & $30$ & $60$ & $120$ & $240$ \\
\midrule
AlpacaFarm & FsfairX & Llama & \textbf{0.985} & \textbf{0.976} & \textbf{0.998} & 1.044 & 1.101 \\
 &  & Gemma & \textbf{0.977} & \textbf{0.962} & \textbf{0.965} & \textbf{0.999} & 1.104 \\
 &  & Mistral & \textbf{0.980} & \textbf{0.960} & \textbf{0.969} & 1.026 & 1.072 \\
 &  & Qwen & \textbf{0.968} & \textbf{0.936} & \textbf{0.928} & \textbf{0.959} & 1.008 \\
\addlinespace
 & RM-Mistral & Llama & \textbf{0.972} & \textbf{0.946} & \textbf{0.948} & \textbf{0.979} & 1.002 \\
 &  & Gemma & \textbf{0.958} & \textbf{0.931} & \textbf{0.930} & \textbf{0.958} & 1.008 \\
 &  & Mistral & \textbf{0.985} & \textbf{0.970} & \textbf{0.991} & 1.009 & 1.037 \\
 &  & Qwen & \textbf{0.964} & \textbf{0.939} & \textbf{0.921} & \textbf{0.957} & \textbf{0.997} \\
\addlinespace
 & ArmoRM & Llama & \textbf{0.994} & \textbf{0.999} & 1.024 & 1.051 & 1.124 \\
 &  & Gemma & \textbf{0.974} & \textbf{0.947} & \textbf{0.963} & 1.026 & 1.110 \\
 &  & Mistral & \textbf{0.988} & \textbf{0.972} & \textbf{0.973} & \textbf{0.973} & 1.008 \\
 &  & Qwen & \textbf{0.966} & \textbf{0.951} & \textbf{0.957} & 1.015 & 1.135 \\
\addlinespace
HH-RLHF & FsfairX & Llama & \textbf{0.994} & \textbf{0.993} & 1.023 & 1.044 & 1.077 \\
 &  & Gemma & \textbf{0.963} & \textbf{0.931} & \textbf{0.924} & \textbf{0.966} & 1.039 \\
 &  & Mistral & \textbf{0.978} & \textbf{0.940} & \textbf{0.909} & \textbf{0.898} & \textbf{0.886} \\
 &  & Qwen & \textbf{0.967} & \textbf{0.933} & \textbf{0.915} & \textbf{0.922} & 1.015 \\
\addlinespace
 & RM-Mistral & Llama & \textbf{0.979} & \textbf{0.949} & \textbf{0.929} & \textbf{0.947} & \textbf{0.997} \\
 &  & Gemma & \textbf{0.960} & \textbf{0.919} & \textbf{0.909} & \textbf{0.964} & \textbf{1.000} \\
 &  & Mistral & \textbf{0.964} & \textbf{0.920} & \textbf{0.912} & \textbf{0.927} & \textbf{0.949} \\
 &  & Qwen & \textbf{0.978} & \textbf{0.932} & \textbf{0.894} & \textbf{0.874} & \textbf{0.925} \\
\addlinespace
 & ArmoRM & Llama & \textbf{0.987} & \textbf{0.977} & \textbf{0.990} & 1.019 & 1.040 \\
 &  & Gemma & \textbf{0.972} & \textbf{0.943} & \textbf{0.938} & \textbf{0.965} & \textbf{0.983} \\
 &  & Mistral & \textbf{0.977} & \textbf{0.953} & \textbf{0.944} & \textbf{0.957} & \textbf{1.000} \\
 &  & Qwen & \textbf{0.980} & \textbf{0.970} & \textbf{0.959} & \textbf{0.965} & \textbf{0.989} \\
\bottomrule
\end{tabular}%

\end{table}

\subsection{Rescaling rewards for the shifted-exponential estimator}
\label{apx:reward_scale}

Before fitting the shifted-exponential estimator \eqref{eq:shifted_exp}, we divide every reward by a single constant, the reward scale
\begin{equation*}
  s = \operatorname{median}_x \operatorname{std}(r \mid x),
\end{equation*}
so that the median within-prompt standard deviation is 1. The scale is estimated once offline, from a sample of prompts with several scored responses each: take each prompt's sample standard deviation, then the median across prompts. In our experiments each configuration uses the $s$ computed from its full pool of $100$ prompts and $960$ responses, listed in \cref{tab:reward_scale_effect}. Only the CDF estimate is affected -- the reported reward and the oracle policy are rank transforms of the pool, which any rescaling leaves unchanged.

\paragraph{Why rescaling is needed.} The estimator is fitted to the \emph{exponentiated} reward, so without the rescaling (\eqref{eq:shifted_exp} applied to $r$ rather than $r/s$) its fitted tail shape depends on the units in which rewards are reported, and not only on the shape of the reward distribution. Writing a first-batch reward as $r = \hat q_\alpha + s z$ with $\hat q_\alpha$ now the $\alpha$-quantile of the rewards $r$ and $z$ of order one, the quantity the unrescaled fit treats as exponential is $e^r - \theta = \theta(e^{sz} - 1)$. When $s \approx 1$ this grows faster than linearly in $z$, and the implied upper tail of $r$ decays doubly exponentially. When $s \ll 1$, $e^{sz} - 1 \approx sz$ and the fit degenerates to an exponential tail on $r$ itself, which is far heavier than the reward distributions we observe; $\hat F_x(R_1)$ is then biased low, $\EI(\hat F_x(R_1), N_2)$ is overstated, and the policy draws a second batch more often than it should. Dividing by $s$ replaces $e^{sz}$ by $e^{z}$, whatever the reward model's units.

\Cref{tab:reward_scale_effect} shows this in practice, for the estimate itself and after debiasing. In the 16 configurations that use FsfairX and RM-Mistral, $s$ lies between $0.63$ and $0.97$: their rewards already have a median within-prompt standard deviation close to 1, and rescaling raises the estimate's RMSE by at most $0.25 \times 10^{-2}$ and changes the debiased RMSE little. ArmoRM has $s$ between $0.0053$ and $0.0099$, and without rescaling its fit fails as described above: at $N_1 = 80$ the estimate's RMSE is more than one and a half times that of the estimate fitted to rescaled rewards, in every ArmoRM configuration.

Debiasing removes most of this difference. After debiasing, the RMSE without rescaling is within $8\%$ of the RMSE with it in every configuration, ArmoRM's included; for ArmoRM, rescaling still lowers it at $N_1 = 20$ in every configuration. Since the regressor of the predicted Beta removes the bias of its inputs (\cref{apx:cdf_estimation}), rescaling matters much less for our method than for the estimate plugged in directly. We keep it, since it costs little (below) and lowers the debiased RMSE at small $N_1$.

\begin{table}[t]
\caption{The reward scale $s$ of each of the 24 configurations of \cref{sec:experiments:setup}, and the RMSE ($\times 10^{-2}$) of the shifted-exponential estimate $\hat F_x(R_1)$ fitted to the raw rewards $r$ and to the rescaled rewards $r/s$, as the estimate itself (est.) and after debiasing (deb.), at first-batch sizes $N_1 = 20$ and $80$, measured as in \cref{tab:reward_scale_kappa}. $s$ is the median within-prompt standard deviation over the full pool of 100 prompts and 960 responses. Debiasing is by binned means, each half of a configuration's prompts mapped by the map fitted on the other half.}
\label{tab:reward_scale_effect}
\centering
\footnotesize
\resizebox{\ifdim\width>\linewidth\linewidth\else\width\fi}{!}{%
\begin{tabular}{lllrrrrrrrrr}
\toprule
 &  &  &  & \multicolumn{4}{c}{$N_1 = 20$} & \multicolumn{4}{c}{$N_1 = 80$} \\
\cmidrule(lr){5-8} \cmidrule(lr){9-12}
dataset & RM & LM & $s$ & est., $r$ & est., $r/s$ & deb., $r$ & deb., $r/s$ & est., $r$ & est., $r/s$ & deb., $r$ & deb., $r/s$ \\
\midrule
AlpacaFarm & FsfairX & Llama & 0.968 & 4.24 & 4.26 & 3.95 & 3.95 & 1.13 & 1.13 & 1.01 & 1.01 \\
 &  & Gemma & 0.627 & 4.51 & 4.75 & 4.35 & 4.25 & 1.30 & 1.41 & 1.22 & 1.21 \\
 &  & Mistral & 0.840 & 4.21 & 4.34 & 3.82 & 3.79 & 1.04 & 1.12 & 0.95 & 0.95 \\
 &  & Qwen & 0.818 & 4.27 & 4.37 & 3.98 & 3.95 & 1.18 & 1.20 & 1.04 & 1.04 \\
\addlinespace
 & RM-Mistral & Llama & 0.938 & 4.20 & 4.25 & 3.81 & 3.80 & 1.08 & 1.11 & 0.95 & 0.95 \\
 &  & Gemma & 0.670 & 4.10 & 4.34 & 3.82 & 3.75 & 1.16 & 1.28 & 0.97 & 0.97 \\
 &  & Mistral & 0.832 & 4.30 & 4.44 & 3.84 & 3.82 & 1.09 & 1.19 & 0.95 & 0.96 \\
 &  & Qwen & 0.871 & 4.25 & 4.34 & 3.87 & 3.85 & 1.19 & 1.23 & 1.01 & 1.01 \\
\addlinespace
 & ArmoRM & Llama & 0.00992 & 4.23 & 4.18 & 4.08 & 3.93 & 2.65 & 1.07 & 1.02 & 1.02 \\
 &  & Gemma & 0.00666 & 4.55 & 4.57 & 4.48 & 4.18 & 2.73 & 1.57 & 1.25 & 1.18 \\
 &  & Mistral & 0.00924 & 4.22 & 4.22 & 4.10 & 3.90 & 2.49 & 1.07 & 1.00 & 0.98 \\
 &  & Qwen & 0.00838 & 4.29 & 4.24 & 4.16 & 3.94 & 2.63 & 1.26 & 1.07 & 1.03 \\
\addlinespace
HH-RLHF & FsfairX & Llama & 0.801 & 4.16 & 4.27 & 3.98 & 3.96 & 1.21 & 1.16 & 0.99 & 0.99 \\
 &  & Gemma & 0.673 & 4.11 & 4.33 & 3.89 & 3.83 & 1.04 & 1.08 & 0.94 & 0.93 \\
 &  & Mistral & 0.803 & 4.20 & 4.37 & 3.78 & 3.74 & 0.99 & 1.12 & 0.90 & 0.90 \\
 &  & Qwen & 0.748 & 4.14 & 4.32 & 3.85 & 3.81 & 0.96 & 1.06 & 0.91 & 0.91 \\
\addlinespace
 & RM-Mistral & Llama & 0.798 & 4.13 & 4.27 & 3.86 & 3.83 & 0.99 & 1.06 & 0.92 & 0.92 \\
 &  & Gemma & 0.680 & 4.07 & 4.31 & 3.82 & 3.75 & 0.98 & 1.11 & 0.93 & 0.91 \\
 &  & Mistral & 0.739 & 4.21 & 4.46 & 3.77 & 3.71 & 0.98 & 1.18 & 0.87 & 0.86 \\
 &  & Qwen & 0.795 & 4.11 & 4.25 & 3.81 & 3.78 & 0.94 & 1.04 & 0.89 & 0.89 \\
\addlinespace
 & ArmoRM & Llama & 0.00661 & 4.36 & 4.20 & 4.16 & 3.97 & 2.88 & 1.07 & 1.03 & 0.99 \\
 &  & Gemma & 0.00533 & 4.27 & 4.18 & 4.12 & 3.87 & 2.79 & 1.09 & 1.03 & 0.98 \\
 &  & Mistral & 0.00654 & 4.33 & 4.26 & 4.16 & 3.93 & 2.87 & 1.08 & 1.05 & 0.97 \\
 &  & Qwen & 0.00654 & 4.31 & 4.18 & 4.11 & 3.97 & 2.79 & 1.13 & 1.02 & 1.02 \\
\bottomrule
\end{tabular}
}%

\end{table}

\paragraph{Choice of target spread.} Targeting unit spread is a choice, and \cref{tab:reward_scale_kappa} sweeps a factor $\kappa$ around it, replacing $s$ by $s/\kappa$ in \eqref{eq:shifted_exp}, so that the estimator is fitted to $e^{\kappa r/s}$ and $\kappa = 1$ is the rescaling above. The best $\kappa$ drifts upward with $N_1$, from $0.5$ at $N_1 \leq 20$ to $1$ at $N_1 \geq 160$: a smaller $\kappa$ gives a heavier fitted tail, which lowers $\hat F_x(R_1)$ and offsets its overestimate at small $N_1$, while at large $N_1$ the overestimate is small (\cref{tab:cdf_estimator_rmse}). No single value is best everywhere; $\kappa = 1$ is the best at $N_1 \geq 160$, within $0.04 \times 10^{-2}$ of the best at $N_1 = 80$, within $0.14 \times 10^{-2}$ at $N_1 = 40$, and within $0.3 \times 10^{-2}$ at the smallest first batches, so we use $\kappa = 1$ throughout and introduce no tuned quantity. After debiasing, the RMSE varies with $\kappa$ by less than $4\%$ at every $N_1$, so the choice matters even less for the input of the predicted Beta.

\begin{table}[t]
\caption{RMSE of the shifted-exponential estimate $\hat F_x(R_1)$, and of the same estimate after debiasing, when rewards are scaled to $\kappa$ times unit median within-prompt spread, averaged over the 24 configurations of \cref{sec:experiments:setup}. The paper uses $\kappa = 1$. Averaged over 100 prompts and 300 Monte Carlo draws per prompt; best per column in bold. Stage-1 batches are drawn from the full pool and scored against the normalized reward rank of \cref{apx:quality_score}, as in Table~\ref{tab:estimate_accuracy}, not against the held-out empirical CDF of Table~\ref{tab:cdf_estimator_rmse}, so the values differ slightly from that table's. Debiasing is by binned means, each half of a configuration's prompts mapped by the map fitted on the other half.}
\label{tab:reward_scale_kappa}
\centering
\small
\begin{tabular}{lrrrrrr}
\toprule
 & \multicolumn{6}{c}{RMSE of CDF ($\times 10^{-2}$)} \\
\cmidrule(lr){2-7}
$\kappa$ & $N_1 = 10$ & $20$ & $40$ & $80$ & $160$ & $240$ \\
\midrule
$0.5$ & \textbf{8.27} & \textbf{4.04} & 2.05 & 1.23 & 0.92 & 0.81 \\
$0.7$ & 8.37 & 4.12 & \textbf{2.04} & \textbf{1.11} & 0.74 & 0.62 \\
$0.85$ & 8.45 & 4.21 & 2.09 & 1.11 & 0.69 & 0.56 \\
$1$ & 8.53 & 4.31 & 2.17 & 1.15 & \textbf{0.69} & \textbf{0.54} \\
$1.2$ & 8.65 & 4.45 & 2.30 & 1.23 & 0.72 & 0.55 \\
$1.5$ & 8.83 & 4.67 & 2.49 & 1.35 & 0.77 & 0.57 \\
$2$ & 9.11 & 5.00 & 2.75 & 1.50 & 0.84 & 0.60 \\
\midrule
 & \multicolumn{6}{c}{RMSE of CDF after debiasing ($\times 10^{-2}$)} \\
\cmidrule(lr){2-7}
$\kappa$ & $N_1 = 10$ & $20$ & $40$ & $80$ & $160$ & $240$ \\
\midrule
$0.5$ & 7.75 & 3.96 & 1.97 & 0.98 & \textbf{0.49} & \textbf{0.33} \\
$0.7$ & 7.70 & 3.91 & 1.96 & \textbf{0.97} & 0.50 & 0.33 \\
$0.85$ & 7.67 & 3.89 & 1.95 & 0.98 & 0.50 & 0.34 \\
$1$ & 7.65 & 3.87 & 1.94 & 0.98 & 0.50 & 0.34 \\
$1.2$ & 7.62 & 3.85 & 1.94 & 0.98 & 0.50 & 0.34 \\
$1.5$ & \textbf{7.58} & 3.84 & \textbf{1.94} & 0.99 & 0.51 & 0.34 \\
$2$ & 7.58 & \textbf{3.82} & 1.94 & 0.99 & 0.51 & 0.34 \\
\bottomrule
\end{tabular}%

\end{table}

\paragraph{Cost of the rescaling.} Estimating $s$ needs no training and no additional model, and uses the same kind of offline sample a practitioner already needs in order to choose $(\lambda_{\text{gen}}, \lambda_{\text{lat}})$ (\cref{apx:lambda-selection}). Only its rough magnitude matters, so a small sample suffices (\cref{tab:reward_scale_sample_size}). From $50$ prompts with $20$ responses each -- a thousand generations, once -- the estimate falls between $0.81$ and $1.10$ times its full-pool value $90\%$ of the time. Dividing by such an estimate is equivalent to taking $\kappa$ between roughly $0.9$ and $1.2$ in \cref{tab:reward_scale_kappa}, which changes the RMSE by at most $0.15 \times 10^{-2}$. Small samples understate $s$, since the standard deviation of a few draws is biased low, but the bias is modest from $n = 20$ on.

\begin{table}[t]
\caption{Estimating $s$ offline from $m$ prompts with $n$ scored responses each: the estimate as a ratio to its value on the full pool of 100 prompts and 960 responses, as median [5th, 95th percentile] over 400 resamples, for Llama on AlpacaFarm scored by FsfairX.}
\label{tab:reward_scale_sample_size}
\centering
\small
\begin{tabular}{lrrrr}
\toprule
prompts & $n = 5$ & $n = 10$ & $n = 20$ & $n = 50$ \\
\midrule
$m = 10$ & 0.83 [0.55, 1.30] & 0.90 [0.64, 1.29] & 0.94 [0.65, 1.34] & 0.99 [0.71, 1.36] \\
$m = 25$ & 0.82 [0.64, 1.06] & 0.89 [0.72, 1.17] & 0.92 [0.74, 1.21] & 0.98 [0.78, 1.20] \\
$m = 50$ & 0.83 [0.69, 0.99] & 0.90 [0.75, 1.06] & 0.95 [0.81, 1.10] & 0.97 [0.84, 1.11] \\
\bottomrule
\end{tabular}%

\end{table}

\paragraph{Scale-invariant alternatives.} The scale dependence comes from fixing the exponent in $e^{r}$ rather than fitting it. A Gompertz upper tail, with survival $\exp(-\eta(e^{b y} - 1))$ with $y \geq 0$ the reward minus the first batch's median and both parameters estimated by maximum likelihood, contains our estimator as the case $b = 1$ and is exactly invariant to any affine rescaling of the reward; it is slightly more accurate than ours at large first batches but less accurate at small ones, since a handful of exceedances estimate two parameters poorly. A generalized-Pareto tail is likewise invariant and less accurate at every $N_1$ we tried. We therefore keep the one-parameter fit and set its scale from offline data.

\section{Estimating the expected improvement via regression}
\label{apx:calibration}

\Cref{sec:cdf-computation} estimates the expected improvement by the predicted Beta. This appendix compares four estimates $\hat m(s_1, N_2)$ of the expected improvement $\EI(F_x(R_1), N_2)$, each differing from the one before in one respect:
\begin{itemize}
  \item \emph{Plug-in}: $\hat m(s_1, N_2) = \EI(\hat F_x(R_1), N_2)$, with $\hat F_x$ the shifted exponential of \cref{apx:cdf_estimation}.
  \item \emph{Debiased}: $\hat m(s_1, N_2) = \EI(g(\hat F_x(R_1), N_1), N_2)$, where $g(\hat F_x(R_1), N_1)$ estimates $\E[F_x(R_1) \mid \hat F_x(R_1), N_1]$ by binned means: for each $N_1$, the mean of $F_x(R_1)$ in each of $25$ equal-count bins of $\hat F_x(R_1)$, interpolated linearly between bins.
  \item \emph{Debiased, more inputs}: $\hat m(s_1, N_2) = \EI(\E[F_x(R_1) \mid \phi], N_2)$, with the conditional mean predicted by a gradient-boosted regressor from inputs $\phi$ that summarize the first batch: $N_1$, shifted-exponential estimates of $F_x(R_1)$ at three tail levels, and the positions of a few quantiles of the first batch's rewards (defined below).
  \item \emph{Predicted Beta} (ours): $\hat m(s_1, N_2) = \E[\EI(F_x(R_1), N_2) \mid \phi]$, computed in closed form from a Beta distribution predicted for $F_x(R_1) \mid \phi$, from the same inputs $\phi$ (\cref{sec:cdf-computation}).
\end{itemize}
The last three, the \emph{regression-based estimates}, are each fitted offline on first batches from one half of a configuration's $100$ prompts, the $50$ \emph{fitting prompts}, and applied to the other half, and vice versa, and each configuration has its own fits. $F_x(R_1)$ is replaced by the normalized rank of $R_1$ among the prompt's responses (\cref{apx:quality_score}), both as the target of the fits and in scoring them.

\begin{figure}[htbp]
  \centering
  \includegraphics[width=\textwidth,height=0.8\textheight,keepaspectratio]{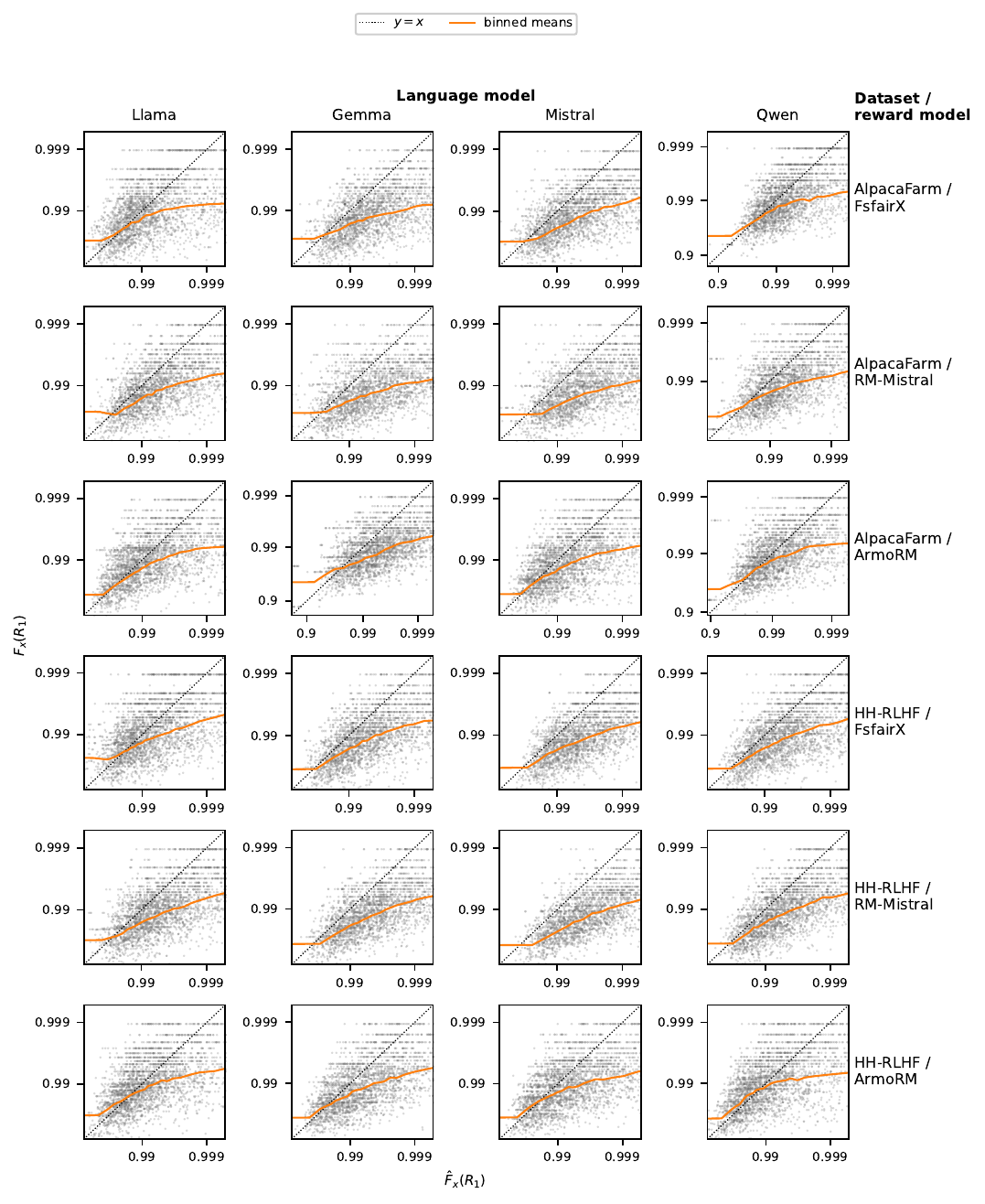}
  \caption{The estimated CDF value $\hat F_x(R_1)$ of the best of $N_1 = 80$ Stage-1 rewards (shifted exponential) against its normalized rank $F_x(R_1)$, in each of the 24 configurations of \cref{sec:experiments:setup} (columns: language models; rows: reward model and dataset). Both axes use a logit scale, $p \mapsto \log(p/(1-p))$, which spreads out values near $1$. Grey points: $3000$ Stage-1 batches sampled at random from all $100$ of the configuration's prompts. Orange: the binned-means map $g(\cdot, 80)$ of the debiased estimate, fitted on one half of those prompts. Dotted: the identity. $g(\cdot, 80)$ falls below the identity at a value between $0.96$ and $0.99$, depending on the configuration: when $\hat F_x(R_1)$ is high, it overestimates $F_x(R_1)$.}
  \label{fig:calibration_scatter}
\end{figure}

\paragraph{Visualizing the bias of $\hat F_x(R_1)$.} \Cref{fig:calibration_scatter} plots $\hat F_x(R_1)$ against $F_x(R_1)$ in each configuration. Their conditional mean is increasing but flatter than the identity: when $\hat F_x(R_1)$ is high, it overestimates $F_x(R_1)$, so the plug-in underestimates the expected improvement. The orange curve is the map $g$ of the debiased estimate.

\paragraph{Fitting procedure of the debiased estimate with more inputs and the predicted Beta.} Both estimates use the same fitted regressor for the mean of $w = N_1(1 - F_x(R_1))$; the predicted Beta adds a second regressor for its variance. The regressors are scikit-learn's histogram gradient boosting \citep{pedregosa2011scikit}. Online, a first batch needs the three shifted-exponential fits, its quantiles and the regressors' predictions, which take negligible time compared to generating a batch.
\begin{itemize}
  \item \emph{Offline data.} For every $N_1 \in \{1, \ldots, 40\} \cup \{50, 60, 80, 100, 120, 160, 240, 320\}$, $400$ first batches of $N_1$ responses are drawn without replacement from each fitting prompt's pool, each with its normalized rank $F_x(R_1)$. Each regressor is one fit over every $N_1$, which is one of its inputs.
  \item \emph{Target.} The mean regressor predicts $w$ rather than $F_x(R_1)$. $F_x(R_1)$ is the maximum of $N_1$ independent $\Unif[0,1]$ values, so $P(w > t) = (1 - t/N_1)^{N_1} \approx e^{-t}$: $w$ is close to an exponential distribution with mean $1$ whatever $N_1$ is, whereas the mean of $1 - F_x(R_1)$, $1/(N_1+1)$, varies more than a hundredfold over the batch sizes. With $w$ as the target, the regressor has to learn only how the first batch's rewards move $w$ from this common reference, not the dependence on $N_1$ known in advance.
  \item \emph{Inputs.} The inputs $\phi$ of a first batch are:
    \begin{itemize}
      \item $\log N_1$, and whether $N_1$ is odd: the shifted exponential's bias alternates between odd and even $N_1$, since a sample of odd size has a reward at its median;
      \item for each tail level $\alpha \in \{1/4, 1/2, 3/4\}$, the rescaled estimate $\hat w_\alpha = N_1(1 - \hat F_{x,\alpha}(R_1))$ (where $\hat F_{x,\alpha}(R_1)$ is the shifted exponential estimator (\eqref{eq:shifted_exp} in \cref{apx:cdf_estimation}) estimated from the rewards above the empirical $\alpha$-quantile $q_\alpha$); we also include a feature indicating whether it fell back to the constant $\hat F_{x,\alpha}(R_1) = N_1/(N_1+1)$, which it does when fewer than five rewards exceed the $\alpha$-quantile (always when $N_1 < 10$ at $\alpha = 1/2$);
      \item the positions $(q_\alpha - q_{0.5})/(R_1 - q_{0.5})$ of the first batch's $0$, $0.1$, $0.25$, $0.75$ and $0.9$ quantiles $q_\alpha$ and of its second-largest reward, relative to its median $q_{0.5}$ and its maximum $R_1$.
    \end{itemize}
    $\phi$ also holds the squares of $\log(\hat w_\alpha + 0.1)$ at $\alpha = 1/4$ and $3/4$, which are functions of the inputs above.
  \item \emph{The mean.} $\hat w(\phi) = e^{\psi_1(\phi)}$ estimates $\E[w \mid \phi]$, with $\psi_1$ a regression model fitted on all $50$ fitting prompts to the targets $y = w$ under the Poisson loss $\ell(\mu, y) = \mu - y \log \mu$, the negative log-likelihood of a Poisson distribution with mean $\mu$ up to terms free of $\mu$. For any nonnegative target $y$, $\E[\ell(\mu, y) \mid \phi]$ is minimized at $\mu = \E[y \mid \phi]$, so this loss fits a conditional mean, and predictions of the form $\mu = e^{\psi(\phi)}$ stay positive. The predicted mean of $F_x(R_1) = 1 - w/N_1$ is $\hat u(\phi) = 1 - \hat w(\phi)/N_1$, clipped to $[0, 1]$. The debiased estimate with more inputs puts it into the closed form: $\hat m(s_1, N_2) = \EI(\hat u(\phi), N_2)$.
  \item \emph{Limiting overfitting.} In each regressor, each tree has at most $31$ leaves of at least $200$ first batches each, and trees are added with learning rate $0.05$. The number of trees, at most $600$, is chosen separately for each regressor by early stopping: the regressor is fitted on $40$ of the $50$ fitting prompts chosen at random, the number of trees minimizing the mean loss on the first batches of the other $10$ is kept, and the regressor is refitted on all $50$ with that number of trees.
\end{itemize}

\paragraph{Additional steps of the predicted Beta.}
\begin{itemize}
  \item \emph{The variance.} $\hat\sigma^2(\phi) = e^{\psi_2(\phi)}$ estimates $\mathrm{Var}(w \mid \phi) = \E[(w - \E[w \mid \phi])^2 \mid \phi]$, with $\psi_2$ fitted under the same loss on all $50$ fitting prompts to the targets $y = (w - \tilde w(\phi))^2$. So that the overfitting of the mean does not shrink these squared residuals, $\tilde w(\phi)$ is computed out of fold: the fitting prompts are split at random into two sets of $25$, and for the first batches of each set, $\tilde w(\phi)$ is the prediction of a mean regressor fitted on the other set, with the number of trees chosen for $\psi_1$.
  \item \emph{The Beta.} $\hat v(\phi) = \hat\sigma^2(\phi)/N_1^2$ is the predicted variance of $F_x(R_1)$. $\mathrm{Beta}(a, b)$ has mean $a/(a+b)$ and variance $\frac{a}{a+b} \cdot \frac{b}{a+b} \cdot \frac{1}{a+b+1}$, so the Beta with mean $\hat u(\phi)$ and variance $\hat v(\phi)$ has $a + b = \hat u(\phi)(1-\hat u(\phi))/\hat v(\phi) - 1$, $a = \hat u(\phi)(a+b)$ and $b = (1-\hat u(\phi))(a+b)$. The sum $a + b$ sets how concentrated the Beta is around its mean. It is floored at $10^{-3}$, since no Beta has mean $\hat u(\phi)$ and variance $\hat v(\phi) \geq \hat u(\phi)(1-\hat u(\phi))$, and $\hat u(\phi)$ is kept strictly inside $(0, 1)$. This Beta gives $\hat m(s_1, N_2) = \E[\EI(F_x(R_1), N_2) \mid \phi]$ in closed form for every $N_2$ (\cref{sec:cdf-computation}).
\end{itemize}

\paragraph{Accuracy.} \Cref{tab:estimate_accuracy} reports each estimate's RMSE $\sqrt{\E[(\hat m(s_1, N_2) - \EI(F_x(R_1), N_2))^2]}$ and its relative bias $\E[\hat m(s_1, N_2)] / \E[\EI(F_x(R_1), N_2)] - 1$, with expectations over first batches and prompts. The rows set $N_2 = N_1$. The relative bias of the debiased estimate with more inputs depends on $N_1$ and $N_2$ mainly through $N_2/N_1$, and moves toward $0$ as $N_2/N_1$ grows (\cref{apx:n2_ratio}); fixing the ratio keeps it comparable across the rows.

Debiasing lowers the RMSE more than the additional inputs do for $N_1 \geq 10$, and less otherwise. The predicted Beta lowers it slightly further for $N_1 \leq 80$; otherwise its RMSE is almost the same as that of the debiased estimate with more inputs. The predicted Beta is the only estimate whose relative bias is small at most $N_1$; the other three underestimate the expected improvement at every $N_1$.

\begin{table}[t]
\caption{Accuracy of the four estimates of the expected improvement $\EI(F_x(R_1), N_2)$, with $N_2 = N_1$, averaged over the 24 configurations of \cref{sec:experiments:setup}. RMSE ($\times 10^{-2}$) against $\EI(F_x(R_1), N_2)$, with $F_x(R_1)$ the normalized rank of $R_1$, lowest in bold; and, for an estimate $\hat m(s_1, N_2)$, the relative bias $\E[\hat m(s_1, N_2)] / \E[\EI(F_x(R_1), N_2)] - 1$, in \%. The regression-based estimates are fitted on half of a configuration's prompts and scored on the other half, both ways round; RMSE is the root of the mean squared error over the 24 configurations.}
\label{tab:estimate_accuracy}
\centering
\small
\resizebox{\ifdim\width>\linewidth\linewidth\else\width\fi}{!}{%
\begin{tabular}{lrrrrrrrr}
\toprule
 & \multicolumn{4}{c}{RMSE ($\times 10^{-2}$)} & \multicolumn{4}{c}{relative bias (\%)} \\
\cmidrule(lr){2-5} \cmidrule(lr){6-9}
$N_1$ & plug-in & debiased & debiased, more inputs & predicted Beta & plug-in & debiased & debiased, more inputs & predicted Beta \\
\midrule
2 & 14.901 & 14.894 & 14.629 & \textbf{14.394} & $-26$ & $-26$ & $-22$ & $+7$ \\
5 & 10.022 & 10.013 & 9.628 & \textbf{9.488} & $-27$ & $-26$ & $-22$ & $+2$ \\
10 & 6.417 & 5.763 & 5.485 & \textbf{5.433} & $-65$ & $-20$ & $-17$ & $+2$ \\
20 & 3.341 & 3.047 & 2.965 & \textbf{2.929} & $-58$ & $-17$ & $-19$ & $-3$ \\
40 & 1.690 & 1.561 & 1.532 & \textbf{1.522} & $-49$ & $-15$ & $-13$ & $+1$ \\
80 & 0.901 & 0.790 & 0.781 & \textbf{0.779} & $-41$ & $-15$ & $-10$ & $+4$ \\
160 & 0.589 & 0.410 & \textbf{0.406} & 0.406 & $-31$ & $-14$ & $-7$ & $+7$ \\
320 & 0.465 & 0.217 & \textbf{0.213} & 0.214 & $-17$ & $-13$ & $-4$ & $+10$ \\
\bottomrule
\end{tabular}
}%

\end{table}

\begin{figure}[htbp]
  \centering
  \includegraphics[width=\textwidth]{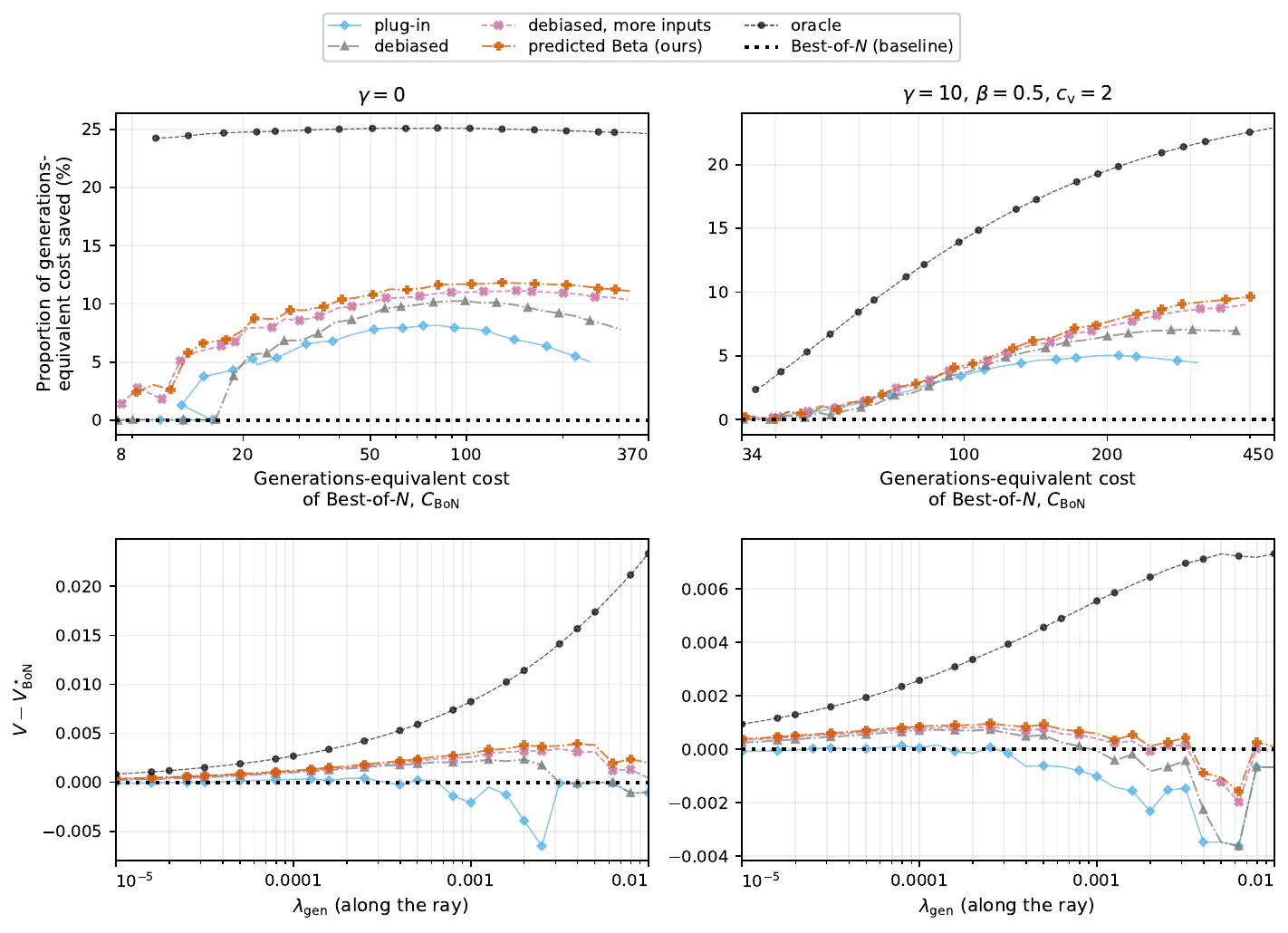}
  \caption{The adaptive batch size, latency-aware policy using each of the four estimates of the expected improvement, batch sizes in $\mathcal{Z} = \{1, \dots, 320\}$, averaged over the 24 configurations of \cref{sec:experiments:setup}. \emph{Left:} no latency cost ($\gamma = 0$). \emph{Right:} the ray $\lambda_{\text{lat}} = 10\lambda_{\text{gen}}$, at $\beta = 0.5$, $c_{\mathrm{v}} = 2$, $\rho = 1$. \emph{Top:} the proportion of generations-equivalent cost $C = \overline N + \gamma\overline T$ saved at matched mean normalized reward rank, $(C_{\text{BoN}} - C_{\text{policy}})/C_{\text{BoN}}$, against Best-of-$N$'s generations-equivalent cost $C_{\text{BoN}}$ in log scale. (Points are joined in order of $\lambda_{\text{gen}}$, so where a policy's mean normalized reward rank is not monotone in $\lambda_{\text{gen}}$, its curve turns back.) \emph{Bottom:} $V - V^\star_{\text{BoN}}$, with $V^\star_{\text{BoN}}$ the value of $V$ of Best-of-$N$ with its $V$-maximizing $N$, against $\lambda_{\text{gen}}$ along the ray (log scale). Light blue diamonds: plug-in; grey triangles: debiased; pink crosses: debiased, more inputs; orange plus signs: predicted Beta (ours); black circles: oracle; dotted black: Best-of-$N$.}
  \label{fig:estimate_ladder}
\end{figure}

\paragraph{Policy performance with each estimate.} \Cref{fig:estimate_ladder} compares the adaptive batch size, latency-aware policy using each of the four estimates with Best-of-$N$ and with the oracle policy, at $\lambda_{\text{lat}} = 0$ and on the ray $\lambda_{\text{lat}} = 10\lambda_{\text{gen}}$ at $\beta = 0.5$ and $c_{\mathrm{v}} = 2$. The top row compares the proportion of generations-equivalent cost saved, where the plug-in always saves cost over Best-of-$N$. The bottom row plots $V - V^\star_{\text{BoN}}$ against $\lambda_{\text{gen}}$, where $V^\star_{\text{BoN}}$ is the value of $V$ of Best-of-$N$ with its $V$-maximizing $N$: at large $\lambda_{\text{gen}}$ the plug-in falls below Best-of-$N$, while the predicted Beta mostly stays above it.
In both views the ordering is the same: the predicted Beta, then the debiased estimate with more inputs close below it, then the debiased estimate, then the plug-in.

\paragraph{Fitting on another configuration.} Each of the 24 configurations has six others that share none of its language model, reward model and dataset: the three other language models crossed with the two other reward models, on the other dataset. \Cref{tab:estimate_transfer} repeats the RMSE of the three regression-based estimates, each fitted instead on half the prompts of each of these six separately, and scored on half of the configuration's own prompts. Fitted this way, each estimate's RMSE is higher than when fitted on the configuration's own prompts, the predicted Beta's by at most a tenth, and at every $N_1$ it is still lower than the RMSE of the plug-in. This compares accuracy only; the policy using an estimate fitted on another configuration was not run.

\begin{table}[t]
\caption{RMSE ($\times 10^{-2}$) against $\EI(F_x(R_1), N_2)$, with $N_2 = N_1$, of the plug-in, which has no fit, and of the three regression-based estimates of the expected improvement, over the 24 configurations of \cref{sec:experiments:setup}. ``Own'': fitted on half of the configuration's prompts and scored on the other half, both ways round. ``Other'': fitted on half of the prompts of another configuration that shares none of the configuration's language model, reward model and dataset, for each of the six such configurations, and scored on half of the configuration's prompts. RMSE is the root of the mean squared error over the 24 configurations, both prompt halves and, for ``other'', the six configurations fitted on.}
\label{tab:estimate_transfer}
\centering
\small
\begin{tabular}{lrrrrrrr}
\toprule
 &  & \multicolumn{2}{c}{debiased} & \multicolumn{2}{c}{debiased, more inputs} & \multicolumn{2}{c}{predicted Beta} \\
\cmidrule(lr){3-4} \cmidrule(lr){5-6} \cmidrule(lr){7-8}
$N_1$ & plug-in & own & other & own & other & own & other \\
\midrule
2 & 14.901 & 14.894 & 14.894 & 14.629 & 14.758 & 14.394 & 14.495 \\
5 & 10.022 & 10.013 & 10.013 & 9.628 & 9.680 & 9.488 & 9.525 \\
10 & 6.417 & 5.763 & 5.772 & 5.485 & 5.543 & 5.433 & 5.497 \\
20 & 3.341 & 3.047 & 3.074 & 2.965 & 3.021 & 2.929 & 2.990 \\
40 & 1.690 & 1.561 & 1.595 & 1.532 & 1.598 & 1.522 & 1.592 \\
80 & 0.901 & 0.790 & 0.823 & 0.781 & 0.835 & 0.779 & 0.835 \\
160 & 0.589 & 0.410 & 0.431 & 0.406 & 0.439 & 0.406 & 0.440 \\
320 & 0.465 & 0.217 & 0.225 & 0.213 & 0.232 & 0.214 & 0.234 \\
\bottomrule
\end{tabular}%

\end{table}

\subsection{How the convexity correction depends on $N_2/N_1$}
\label{apx:n2_ratio}

\Cref{tab:estimate_accuracy,tab:estimate_transfer} vary $N_1$ with $N_2 = N_1$, so that the relative size of the predicted Beta's correction for the convexity of $\EI$ differs across rows approximately only through the uncertainty about $F_x(R_1)$ given the first batch. With $U = F_x(R_1)$ and $w = N_1(1 - U)$, as in the fitting procedure, $\EI(u, N) = (1 - u) + \frac{u^{N+1} - 1}{N+1}$. $w$ is of order one (\cref{apx:calibration}, fitting procedure), so $w/N_1$ is small and $\log(U) = \log(1 - w/N_1) \approx -w/N_1$. Hence, with $\xi \coloneqq N_2/N_1$ and $N_2 + 1 \approx N_2$, $U^{N_2+1} = \exp\!\big((N_2+1) \log(1 - w/N_1)\big) \approx e^{-\xi w}$. Substituting $1 - U = w/N_1$ gives
\begin{equation*}
  \EI(U, N_2) \approx \frac{H(\xi w)}{N_2}, \qquad H(t) \coloneqq t - 1 + e^{-t},
\end{equation*}
and likewise $\EI(\E[U \mid \phi], N_2) \approx H(\xi \E[w \mid \phi])/N_2$. The relative bias given $\phi$ of the debiased estimate with more inputs is therefore
\begin{equation*}
  \frac{\EI(\E[U \mid \phi], N_2)}{\E[\EI(U, N_2) \mid \phi]} - 1 \approx \delta(\xi) \coloneqq \frac{H(\xi \E[w \mid \phi])}{\E[H(\xi w) \mid \phi]} - 1,
\end{equation*}
which depends on $N_1$ and $N_2$ only through $\xi$, given the conditional distribution of $w$, and is at most $0$ by Jensen's inequality.

\emph{The limits.} By Taylor expansion of $H$ at $0$, $H(t) \approx t^2/2$ for small $t$, so as $\xi \to 0$, $\delta(\xi) \to \E[w \mid \phi]^2/\E[w^2 \mid \phi] - 1 = -\mathrm{Var}(w \mid \phi)/\E[w^2 \mid \phi]$. $H(t) \approx t$ for large $t$, so as $\xi \to \infty$, $\delta(\xi) \to \E[w \mid \phi]/\E[w \mid \phi] - 1 = 0$: $\EI(U, N_2)$ is then nearly linear in $U$.

\emph{The increase in $\xi$.} $\delta$ increases in $\xi$ for every conditional distribution of $w$. Let $t = \xi w$ and $\bar t = \xi \E[w \mid \phi]$. By the quotient rule,
\begin{equation*}
  \delta'(\xi) = -\frac{H(\bar t)\, \E[k(t) \mid \phi]}{\xi\, \E[H(t) \mid \phi]^2}, \qquad k(t) \coloneqq t H'(t) - \frac{\bar t H'(\bar t)}{H(\bar t)} H(t),
\end{equation*}
so it suffices to show $\E[k(t) \mid \phi] \leq 0$. Since $k(\bar t) = 0$, the tangent line of $k$ at $\bar t$ is $k'(\bar t)(t - \bar t)$. So it suffices to show that $k$ lies below this line:
\begin{equation*}
  k(t) \leq k'(\bar t)(t - \bar t),
\end{equation*}
since taking expectations of both sides gives $\E[k(t) \mid \phi] \leq 0$. We have $k(0) = k'(0) = 0$, and $k''(t) = (t_0 - t)e^{-t}$ with $t_0 \coloneqq 2 - \bar t H'(\bar t)/H(\bar t)$, so $k$ is strictly convex below $t_0$ and strictly concave above it. Then $0 < t_0 < \bar t$: if $t_0 \leq 0$, $k$ would be strictly concave on $[0, \bar t]$, and $k(0) = k'(0) = 0$ would give $k(\bar t) < 0$; if $t_0 \geq \bar t$, it would be strictly convex there and give $k(\bar t) > 0$. Also $k'(\bar t) < 0$: $k$ increases on $[0, t_0]$ from $k(0) = 0$, so $k(t_0) > 0 = k(\bar t)$, hence $k'$ is negative somewhere in $(t_0, \bar t)$, and $k'$ decreases above $t_0$. The inequality follows: above $t_0$ because $k$ is concave there and $t_0 < \bar t$, and for $t < t_0$ because $k(t) \leq k(t_0) \leq k'(\bar t)(t_0 - \bar t) \leq k'(\bar t)(t - \bar t)$: the first inequality holds since $k$ increases on $[0, t_0]$, as $k'(0) = 0$ and $k'' > 0$ there; the second by the concave case at $t_0$; the third since $k'(\bar t) < 0$.

Holding $N_2/N_1$ fixed across the rows therefore approximately holds the relative size of the correction fixed for a given conditional distribution of $w$, and the rows differ only in the uncertainty about $F_x(R_1)$ given the first batch.

\section{Additional figures and tables of experiment results}

\subsection{Results across latency costs and dependence of latency on batch size}
\label{apx:gec_grid}

\begin{figure}[htbp]
  \centering
  \includegraphics[width=\textwidth]{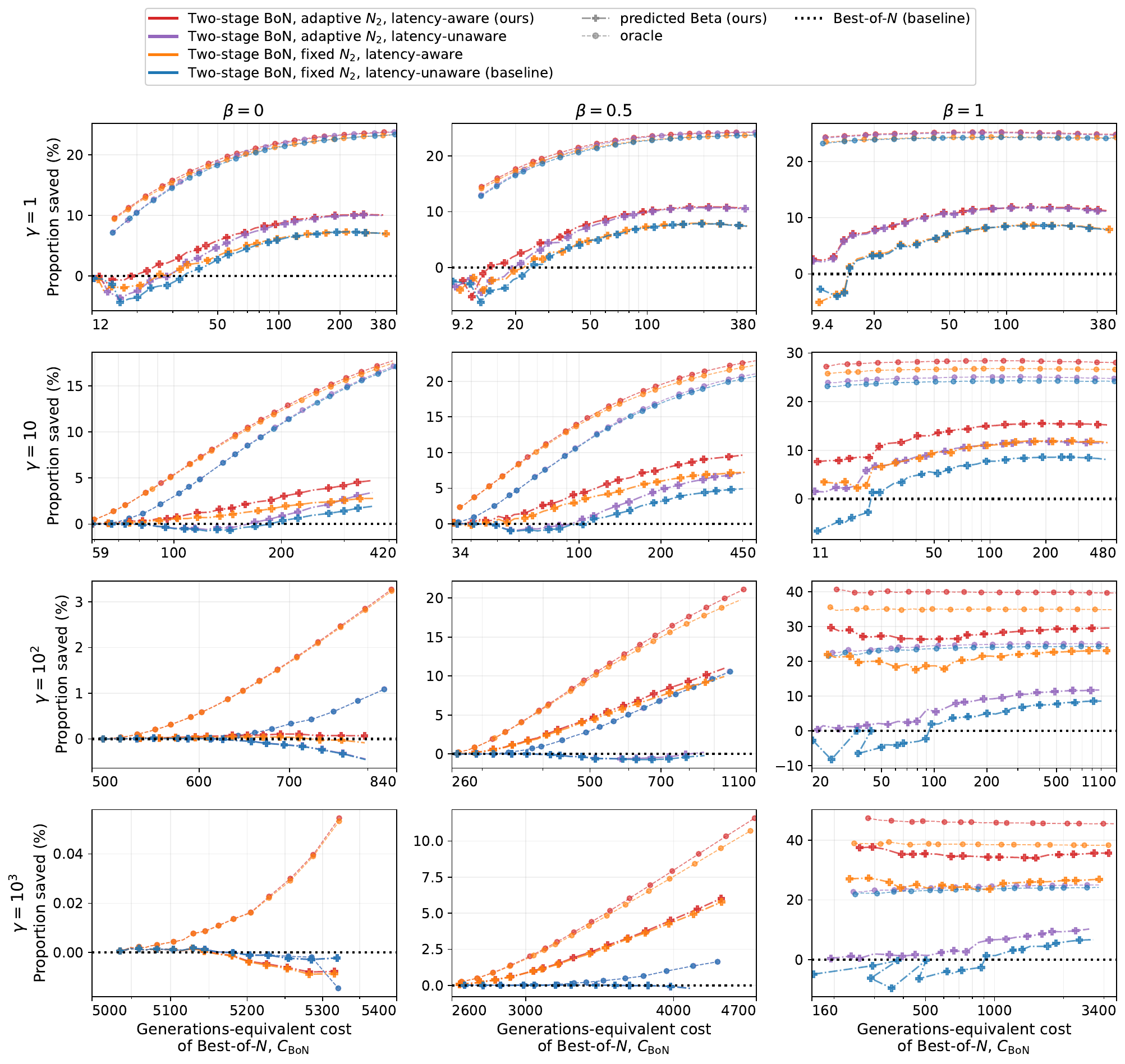}
  \caption{The bottom row of \cref{fig:gec_gamma} at each $\beta \in \{0,0.5,1\}$, also with $\gamma=1$: proportion of generations-equivalent cost $C = \overline N + \gamma\overline T$ saved at matched mean normalized reward rank, $(C_{\text{BoN}} - C_{\text{policy}})/C_{\text{BoN}}$, against Best-of-$N$'s generations-equivalent cost $C_{\text{BoN}}$ (log scale), at $\gamma = 1$, $10$, $10^2$ and $10^3$ (rows) and $\beta = 0$, $0.5$ and $1$ (columns). $c_{\mathrm{v}} = 2$, $\rho = 1$, batch sizes in $\mathcal{Z} = \{1, \dots, 320\}$, averaged over the 24 configurations of \cref{sec:experiments:setup}. Points are joined in order of $\lambda_{\text{gen}}$, so where a policy's mean normalized reward rank is not monotone in $\lambda_{\text{gen}}$, its curve turns back. Colours and line styles as in \cref{fig:gec_gamma}.}
  \label{fig:grid_saved}
\end{figure}

\begin{figure}[htbp]
  \centering
  \includegraphics[width=\textwidth]{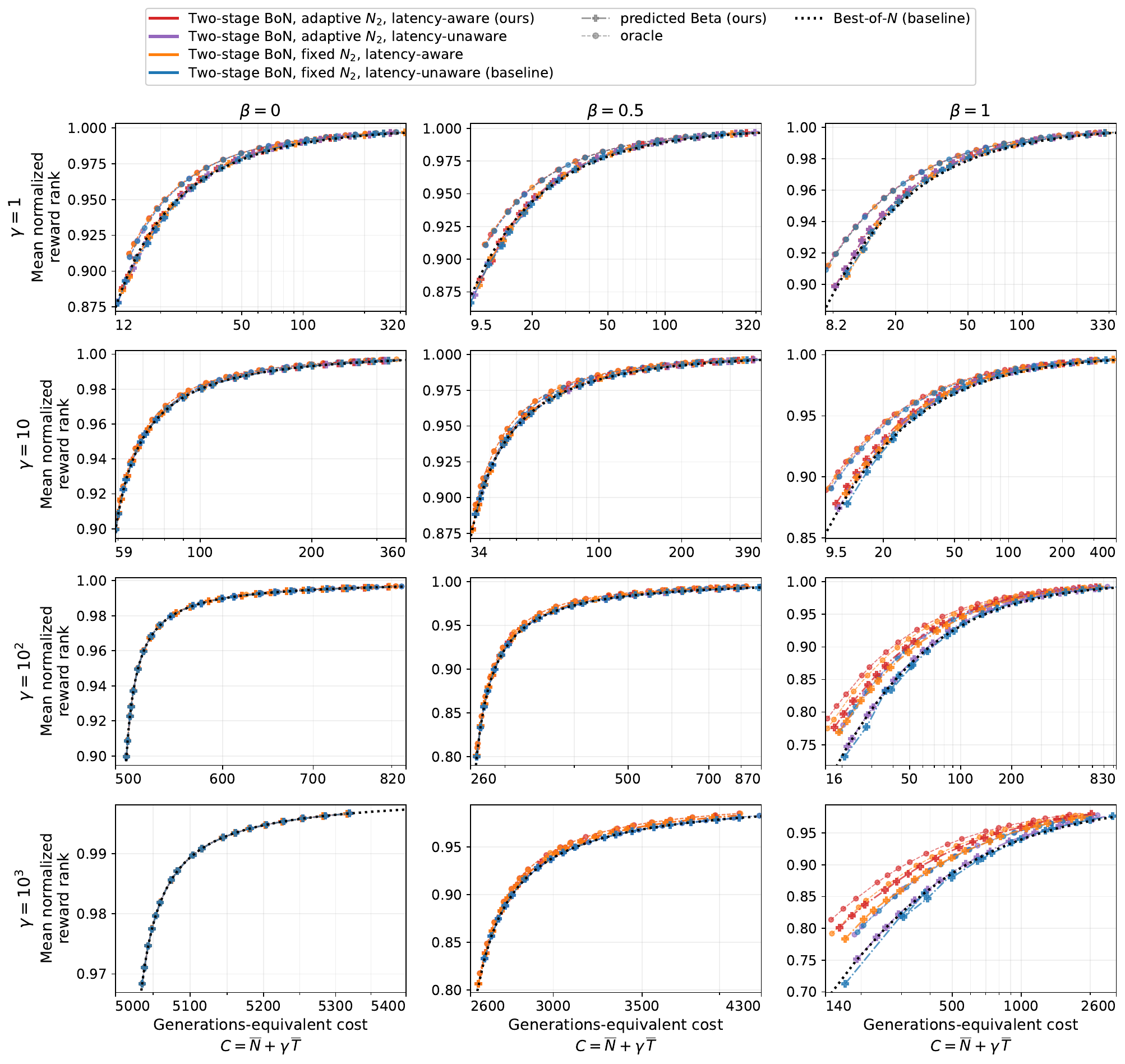}
  \caption{The top row of \cref{fig:gec_gamma} at each $\beta \in \{0,0.5,1\}$, also with $\gamma=1$: mean normalized reward rank against generations-equivalent cost $C = \overline N + \gamma\overline T$ ($\overline N$, $\overline T$: mean number of generations and mean total latency in seconds) (log scale) for the policies and settings of \cref{fig:grid_saved}. Dotted black: Best-of-$N$.}
  \label{fig:grid_reward}
\end{figure}

\Cref{fig:grid_saved,fig:grid_reward} show the curves of \cref{fig:gec_gamma} at $\gamma = 1$, $10$, $10^2$ and $10^3$ (rows) and $\beta = 0$, $0.5$ and $1$ (columns), at $c_{\mathrm{v}} = 2$ and $\rho = 1$: the proportion of generations-equivalent cost saved at Best-of-$N$'s reward, and the mean normalized reward rank, against generations-equivalent cost. The middle column of both, $\beta = 0.5$, is \cref{fig:gec_gamma}.

\subsection{Per-configuration results across language models, reward models, and datasets}
\label{apx:scope}

\begin{figure}[htbp]
  \centering
  \includegraphics[width=\textwidth]{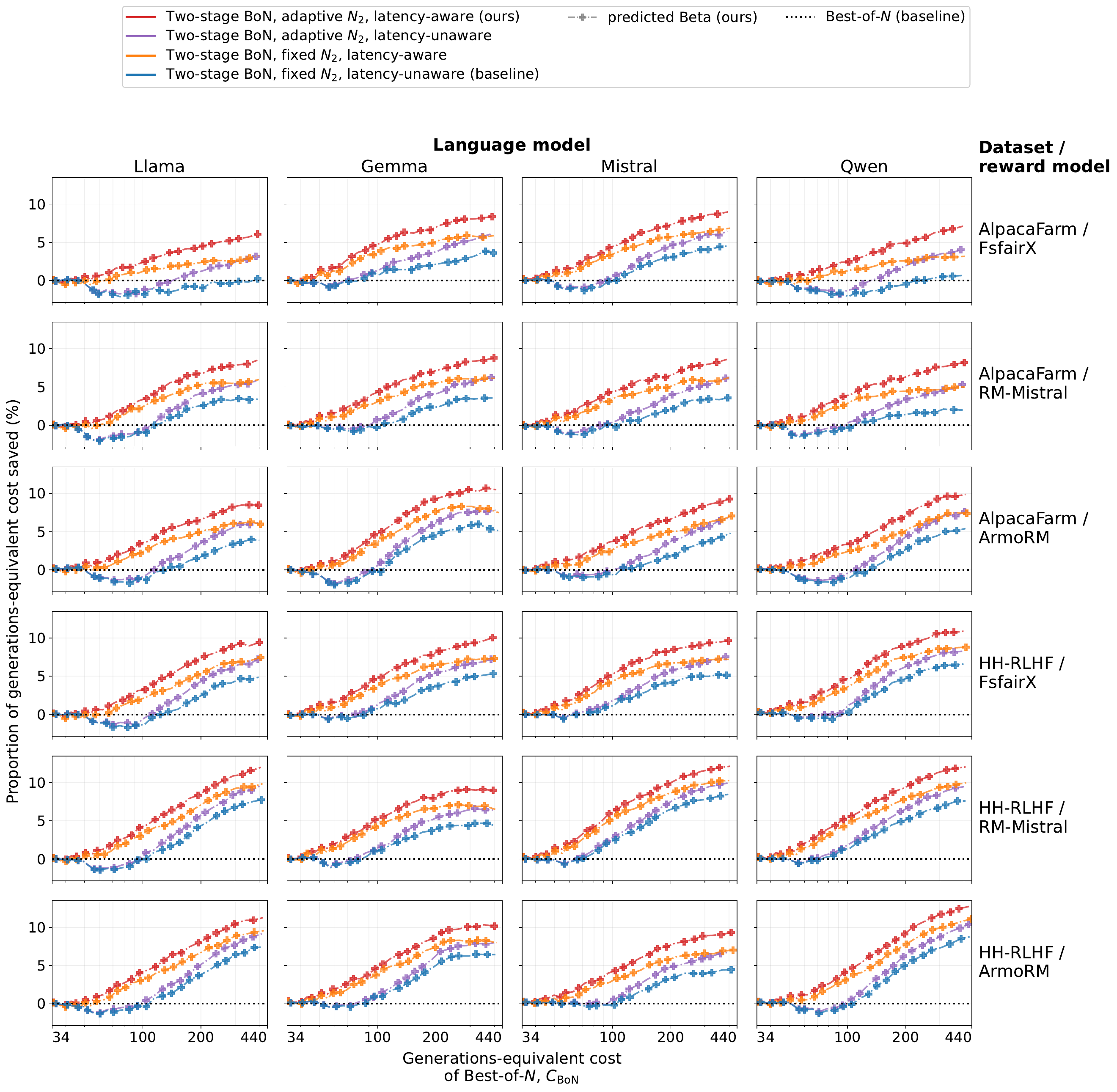}
  \caption{The bottom row of \cref{fig:gec_gamma} at $\gamma = 10$ in each of the 24 configurations of \cref{sec:experiments:setup}: the proportion of generations-equivalent cost saved against Best-of-$N$'s generations-equivalent cost $C_{\text{BoN}}$ (log scale), at $\beta = 0.5$, $c_{\mathrm{v}} = 2$, $\rho = 1$, batch sizes in $\mathcal{Z} = \{1, \dots, 320\}$, with the predicted Beta; each panel uses its configuration's own Best-of-$N$. Columns: language models; rows: reward model and dataset.}
  \label{fig:scope_gec}
\end{figure}

\Cref{sec:experiments} reports results averaged over the 24 configurations (\cref{sec:experiments:setup});
the per-configuration values are shown here.
\Cref{fig:scope_gec} gives the bottom panel of \cref{fig:gec_gamma} at $\gamma = 10$, and \cref{tab:scope_gec} gives the proportion of generations-equivalent cost saved at the reward of Best-of-$N$ with $N = 40$, as in \cref{tab:gec_body} at $\beta = 0.5$ and $\beta = 1$, both with the predicted Beta. We do not report oracle performance here because with the oracle the proportion saved varies little across configurations: the oracle policy sees the rewards only through $F_x(R_1)$, whose distribution does not depend on the configuration.

\begin{table}[t]
\caption{Per-configuration results of \cref{tab:gec_body}: proportion (\%) of generations-equivalent cost saved at the reward of Best-of-$N$ with $N = 40$ in each of the 24 configurations of \cref{sec:experiments:setup} (adaptive batch size policies, $\mathcal{Z} = \{1, \dots, 320\}$, predicted Beta, at $c_{\mathrm{v}} = 2$ and $\rho = 1$). At $\gamma = 0$ the two latency rules are one policy and $\beta$ does not apply.}
\label{tab:scope_gec}
\centering
\footnotesize
\resizebox{\ifdim\width>\linewidth\linewidth\else\width\fi}{!}{%
\begin{tabular}{lllrrrrrrrrr}
\toprule
 &  &  &  & \multicolumn{4}{c}{$\beta = 0.5$} & \multicolumn{4}{c}{$\beta = 1$} \\
\cmidrule(lr){5-8} \cmidrule(lr){9-12}
 &  &  & \multicolumn{1}{c}{$\gamma = 0$} & \multicolumn{2}{c}{$\gamma = 10$} & \multicolumn{2}{c}{$\gamma = 10^2$} & \multicolumn{2}{c}{$\gamma = 10$} & \multicolumn{2}{c}{$\gamma = 10^2$} \\
\cmidrule(lr){4-4} \cmidrule(lr){5-6} \cmidrule(lr){7-8} \cmidrule(lr){9-10} \cmidrule(lr){11-12}
dataset & RM & LM &  & unaware & aware & unaware & aware & unaware & aware & unaware & aware \\
\midrule
AlpacaFarm & FsfairX & Llama & $8.8$ & $-1.9$ & $1.6$ & $-0.2$ & $2.1$ & $8.8$ & $12.8$ & $8.4$ & $27.5$ \\
 &  & Gemma & $11.2$ & $0.3$ & $3.4$ & $0.0$ & $2.9$ & $11.4$ & $15.1$ & $11.1$ & $29.1$ \\
 &  & Mistral & $11.0$ & $-0.7$ & $2.9$ & $-0.3$ & $2.7$ & $11.0$ & $14.9$ & $10.8$ & $28.8$ \\
 &  & Qwen & $8.1$ & $-1.5$ & $1.5$ & $-0.2$ & $2.2$ & $8.1$ & $12.5$ & $8.1$ & $27.3$ \\
\addlinespace
 & RM-Mistral & Llama & $9.3$ & $-1.2$ & $1.8$ & $-0.7$ & $2.2$ & $9.3$ & $13.2$ & $8.7$ & $27.6$ \\
 &  & Gemma & $9.9$ & $-0.6$ & $2.7$ & $-0.2$ & $2.5$ & $10.1$ & $13.7$ & $10.0$ & $28.0$ \\
 &  & Mistral & $10.8$ & $-0.3$ & $2.9$ & $-0.1$ & $2.7$ & $11.0$ & $14.9$ & $10.8$ & $28.8$ \\
 &  & Qwen & $9.9$ & $-0.5$ & $2.7$ & $-0.3$ & $2.6$ & $10.1$ & $13.9$ & $9.9$ & $28.0$ \\
\addlinespace
 & ArmoRM & Llama & $9.6$ & $-1.3$ & $2.0$ & $0.0$ & $2.5$ & $9.7$ & $13.8$ & $9.2$ & $28.1$ \\
 &  & Gemma & $10.7$ & $-1.0$ & $2.6$ & $-0.2$ & $2.6$ & $10.8$ & $14.6$ & $10.5$ & $28.6$ \\
 &  & Mistral & $9.7$ & $-0.7$ & $2.2$ & $0.1$ & $2.5$ & $9.9$ & $13.7$ & $9.3$ & $28.0$ \\
 &  & Qwen & $9.2$ & $-1.4$ & $2.1$ & $-0.4$ & $2.4$ & $9.0$ & $13.1$ & $8.6$ & $27.8$ \\
\addlinespace
HH-RLHF & FsfairX & Llama & $8.5$ & $-1.3$ & $1.7$ & $-0.1$ & $2.3$ & $8.5$ & $12.6$ & $8.1$ & $27.6$ \\
 &  & Gemma & $10.3$ & $-0.1$ & $3.0$ & $0.1$ & $2.8$ & $10.5$ & $14.3$ & $10.4$ & $28.3$ \\
 &  & Mistral & $11.0$ & $0.5$ & $3.5$ & $0.1$ & $2.9$ & $11.1$ & $14.7$ & $11.1$ & $28.2$ \\
 &  & Qwen & $10.4$ & $-0.3$ & $3.0$ & $-0.1$ & $2.8$ & $10.5$ & $14.2$ & $9.9$ & $28.1$ \\
\addlinespace
 & RM-Mistral & Llama & $10.1$ & $-0.8$ & $2.5$ & $-0.3$ & $2.5$ & $10.3$ & $14.2$ & $10.1$ & $28.1$ \\
 &  & Gemma & $11.7$ & $0.0$ & $3.4$ & $-0.2$ & $3.1$ & $12.0$ & $15.7$ & $11.9$ & $29.7$ \\
 &  & Mistral & $11.5$ & $0.9$ & $3.9$ & $-0.1$ & $3.0$ & $11.7$ & $15.2$ & $11.7$ & $28.3$ \\
 &  & Qwen & $11.0$ & $0.4$ & $3.7$ & $0.1$ & $3.0$ & $11.2$ & $14.8$ & $10.6$ & $28.6$ \\
\addlinespace
 & ArmoRM & Llama & $9.9$ & $-0.6$ & $2.4$ & $-0.2$ & $2.6$ & $10.0$ & $13.7$ & $9.6$ & $27.9$ \\
 &  & Gemma & $10.8$ & $0.0$ & $3.1$ & $0.1$ & $2.9$ & $11.0$ & $14.8$ & $10.8$ & $28.7$ \\
 &  & Mistral & $9.5$ & $-0.2$ & $2.8$ & $-0.1$ & $2.7$ & $9.7$ & $13.2$ & $9.4$ & $27.8$ \\
 &  & Qwen & $9.5$ & $-0.7$ & $2.3$ & $0.0$ & $2.7$ & $9.7$ & $13.5$ & $9.2$ & $27.7$ \\
\bottomrule
\end{tabular}
}%

\end{table}

\subsection{Sensitivity to the $T_1$--$T_2$ correlation}
\label{apx:corr_sweep}

Every latency-aware policy compared in \cref{sec:experiments} uses the exact, correct prediction of $\widehat T_2(s_1, N_2)$, and focuses on the perfect-correlation case $\rho = 1$ (\cref{sec:latency_model_results} also compares to $\rho=0.5$).
Here, we systematically explore how the performance depends on $\rho$, sweeping $\rho \in \{0, 0.25, 0.5, 0.75, 1\}$.
We also compare with a latency-aware policy that uses a misspecified latency predictor, predicting under the assumption that $\rho = 1$, estimating $\widehat T_2(s_1,N_2) = T_1 h(N_2)/h(N_1)$ whatever the true $\rho$, which is the other extreme of the latency-unaware policy that assumes $\rho=0$.

\begin{figure}[htbp]
  \centering
  \includegraphics[width=\textwidth]{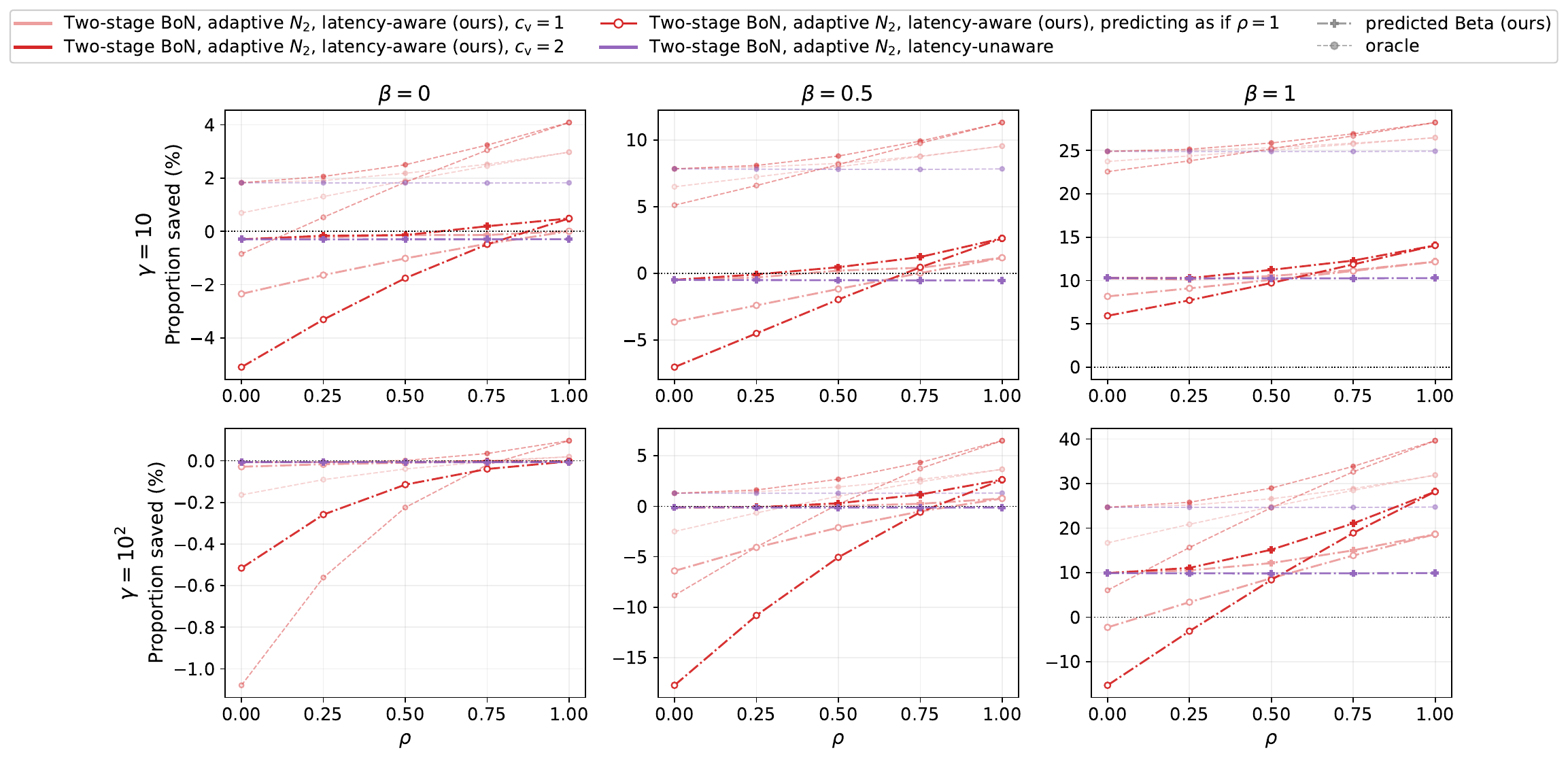}
  \caption{Proportion (\%) of generations-equivalent cost $C = \overline N + \gamma\overline T$ saved at the reward of Best-of-$N$ with $N = 40$, against the correlation $\rho$ of $\log T_1$ and $\log T_2$, at $\gamma = 10$ and $10^2$ (rows) and $\beta = 0$, $0.5$ and $1$ (columns). All policies choose the second batch size adaptively in $\mathcal{Z} = \{1, \dots, 320\}$. Performance is averaged over the 24 configurations of \cref{sec:experiments:setup}.
  Predicting $T_2$ at the true $\rho$ in filled markers, predicting as if $\rho = 1$ in hollow markers; line styles are otherwise the same as in \cref{fig:gec_beta}.
  }
  \label{fig:corr_sweep}
\end{figure}

\Cref{fig:corr_sweep} shows the proportion of generations-equivalent cost saved at the reward of Best-of-$N$ with $N = 40$, as in \cref{tab:gec_body}, against $\rho$, for the adaptive batch size policies at $\gamma \in \{10, 10^2\}$, $\beta \in \{0, 0.5, 1\}$, and $c_{\mathrm{v}} \in \{1, 2\}$.
The latency-aware policy that predicts $T_2$ correctly saves as much as the latency-unaware policy at $\rho = 0$ (in which case $\widehat T_2(s_1, N_2) = \overline T_2(N_2)$), and more the higher $\rho$ is; we also see higher savings with higher $c_{\mathrm{v}}$ and $\beta$ as already shown in \cref{fig:gec_beta}. Predicting $T_2$ as if $\rho = 1$ never saves more than predicting it at the true $\rho$, and at $\rho \leq 0.25$ it saves less than the latency-unaware policy, by more at $c_{\mathrm{v}} = 2$: reading a $T_1$ that says little about $T_2$ as if it determined $T_2$ is worse than not reading it.

\subsection{Heatmaps of the lambda grid}
\label{apx:heatmaps}
\begin{figure}[htbp]
  \centering
  \includegraphics[width=\textwidth]{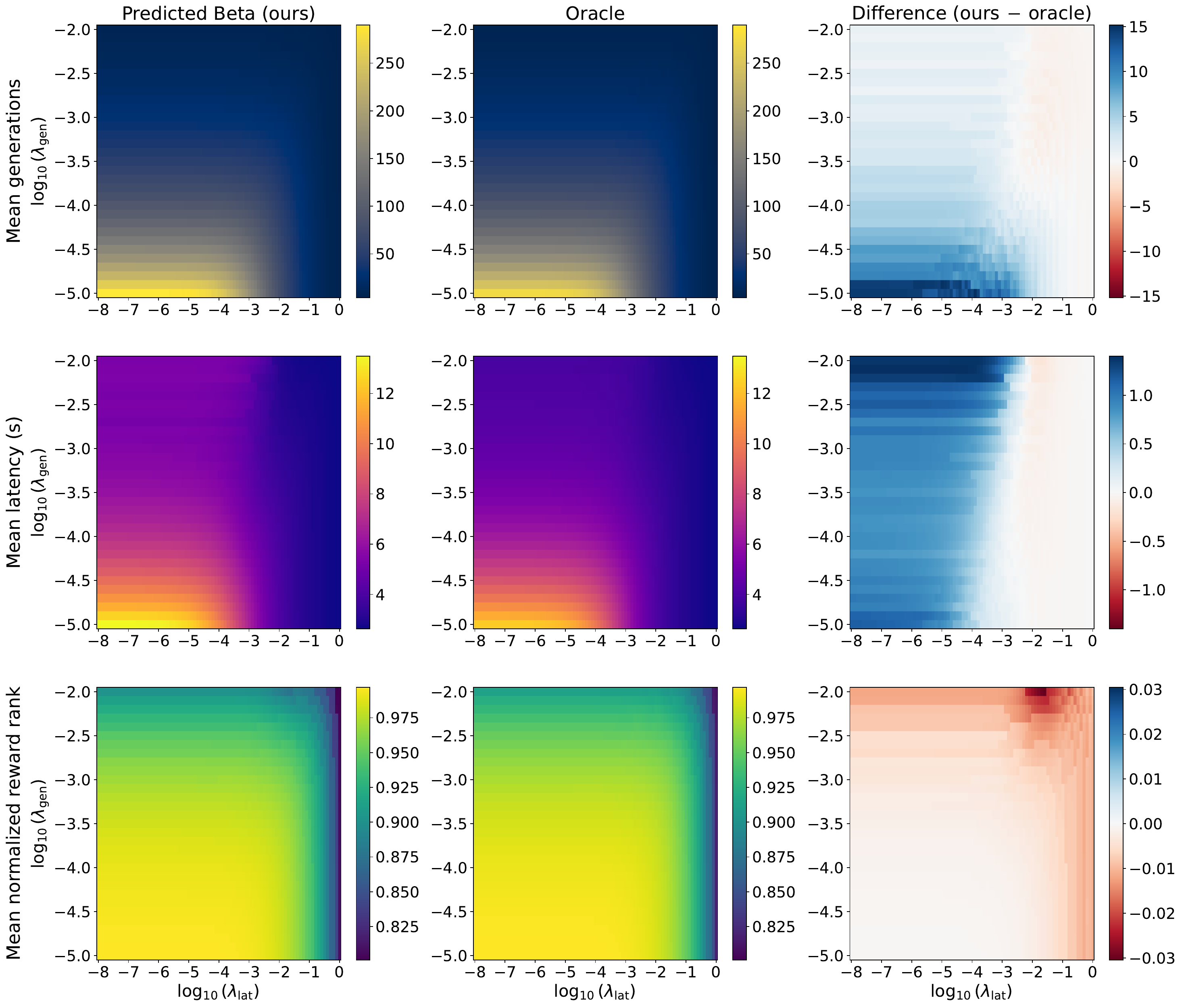}
  \caption{Heatmaps over the grid of the 31 geometrically spaced values of $\lambda_{\text{gen}} \in [10^{-5}, 10^{-2}]$ of \cref{sec:experiments:setup} and 81 geometrically spaced values of $\lambda_{\text{lat}} \in [10^{-8}, 10^{0}]$ for the latency-aware policy with adaptive batch sizes $\mathcal{Z} = \{1, \dots, 320\}$, at $\beta = 0.5$, $c_{\mathrm{v}} = 2$, $\rho = 1$, averaged over the 24 configurations of \cref{sec:experiments:setup}.
    Top row: mean number of generations.
    Middle row: mean latency (s).
    Bottom row: mean normalized reward rank of the returned response.
    Left column: \emph{our method}, with the predicted Beta. Middle column: oracle variant. Right column: ours minus oracle.}
  \label{fig:heatmap}
\end{figure}

\Cref{fig:heatmap} shows, over the grid of the 31 geometrically spaced values of $\lambda_{\text{gen}} \in [10^{-5}, 10^{-2}]$ of \cref{sec:experiments:setup} and 81 geometrically spaced values of $\lambda_{\text{lat}} \in [10^{-8}, 10^{0}]$, the mean normalized reward rank, the mean latency, and the mean number of generations of the latency-aware policy with adaptive batch sizes, at the latency model of \cref{fig:gec_gamma} ($\beta = 0.5$, $c_{\mathrm{v}} = 2$, $\rho = 1$), averaged over the 24 configurations of \cref{sec:experiments:setup}. Each ray of \cref{fig:gec_gamma} is a diagonal of this grid.

\section{Selecting the tradeoff parameters in practice}
\label{apx:lambda-selection}
The two tradeoff parameters $(\lambda_{\text{gen}}, \lambda_{\text{lat}})$ are selected once, offline. Run inference on an initial serving trace or a held-out set, sweep $(\lambda_{\text{gen}}, \lambda_{\text{lat}})$ over a grid as we do above, plot the resulting tradeoff among mean reward, mean latency, and mean generation count, and select the pair that meets the deployment target. Two properties keep this cheap. First, the $\lambda$'s do not need to adapt to load: adaptation to the state of the serving system is automatic through $T_1$, in contrast to the fixed-threshold or fixed-cost rules that many adaptive TTC methods use, which do not respond to the state of the serving system. Second, because the CDF transformation places the decision on the scale-free $[0,1]$ scale of CDF values, $\lambda_{\text{gen}}$ and $\lambda_{\text{lat}}$ should transfer across language models and reward models as long as the serving system -- and hence the latency distribution -- is unchanged; \cref{apx:scope} checks this across the 24 configurations.

\section{Random draws and seeding}
\label{apx:seeding}

Every random draw in the simulations comes from a generator with a fixed seed, so each run reproduces exactly, and the same draws are shared across all the policies being compared (common random numbers). The latencies $T_{1,i}$, and $T_{2,i}$ when $\rho < 1$ (\cref{apx:corr_sweep}), are drawn once per prompt and trial. Trial $i$ draws one uniformly random ordering of the pool $\mathcal{R}_b$: its first $N_1$ entries form the Stage-1 set $A_{b,i}$ and the next $N_{2,i}$ the second batch $B_{b,i}$, which is equivalent to the two draws without replacement of \cref{sec:experiments:setup}. The same orderings index every prompt's pool; since each pool holds independent generations, this leaves each prompt's draws with the same distribution. These latencies and orderings are reused for every policy, every $(\lambda_{\text{gen}}, \lambda_{\text{lat}})$ pair, and both the estimated-EI and oracle variants. The empirical Best-of-$N$ reference curves involve no random draws: they are computed exactly (\cref{apx:quality_score}). Because the second batch continues the first batch's ordering, trial $i$ sees the same rewards whatever $N_1$ and $N_{2,i}$ a policy chooses. A difference between two policies therefore reflects only their decisions, not different random draws, which is why the latency-aware policy at $\rho = 0$ matches the latency-unaware baseline exactly rather than to within Monte Carlo noise. The latency samples used for the offline choice of batch sizes (\cref{apx:optimal-N1}) come from a separately seeded generator, so the chosen $N_1$ and $N_2$ are also reproducible. All 24 configurations of \cref{sec:experiments:setup} use the same seed. The regressor of the predicted Beta (\cref{apx:calibration}) is fitted once for each half of each configuration's prompts, and every run loads that fit; its predictions reproduce exactly on one type of CPU and differ across types only by floating-point rounding of its inputs.

\section{Dataset and model asset details}
\label{apx:assets}

All experiments reuse the pools of generations and reward scores released by \citet{kalayci2025optimal}: for each of 100 prompts from each dataset, 960 responses sampled from each language model and scored by each reward model. We use every combination of the four language models, three reward models, and two datasets below, the 24 configurations of \cref{sec:experiments:setup}, and generate or score no new responses. The three reward models are also those used by \citet{sun2024fast}.

\paragraph{Datasets.} We use prompts from the following datasets, 100 from each:

\begin{itemize}
    \item \textbf{AlpacaFarm}: \url{https://github.com/tatsu-lab/alpaca_farm}, Apache 2.0 License.
    \item \textbf{HH-RLHF}: \url{https://huggingface.co/datasets/Anthropic/hh-rlhf}, MIT License.
\end{itemize}

\paragraph{Language models (LMs).} Responses were generated by the following publicly available language models:

\begin{itemize}
    \item \textbf{Llama-3.1-8B-Instruct}: \url{https://huggingface.co/meta-llama/Llama-3.1-8B-Instruct}, Llama 3.1 Community License.
    \item \textbf{Gemma-2-9B-it}: \url{https://huggingface.co/google/gemma-2-9b-it}, Gemma Terms of Use.
    \item \textbf{Mistral-7B-Instruct-v0.3}: \url{https://huggingface.co/mistralai/Mistral-7B-Instruct-v0.3}, Apache 2.0 License.
    \item \textbf{Qwen2.5-7B-Instruct}: \url{https://huggingface.co/Qwen/Qwen2.5-7B-Instruct}, Apache 2.0 License.
\end{itemize}

\paragraph{Reward models (RMs).} Responses were scored by the following publicly available reward models:

\begin{itemize}
    \item \textbf{FsfairX-LLaMA3-RM-v0.1}: \url{https://huggingface.co/sfairXC/FsfairX-LLaMA3-RM-v0.1}, CC-BY-NC-4.0 License.
    \item \textbf{RM-Mistral-7B}: \url{https://huggingface.co/weqweasdas/RM-Mistral-7B}; no license is stated on the model card.
    \item \textbf{ArmoRM-Llama3-8B-v0.1}: \url{https://huggingface.co/RLHFlow/ArmoRM-Llama3-8B-v0.1}, Llama 3 Community License.
\end{itemize}

For all models and datasets, we follow their licensing terms and acknowledge the original sources.

\section{Compute infrastructure}
\label{apx:compute}

All experiments are simulations over precomputed rewards and run on an internal university CPU cluster; no GPU is needed. A run fixes one configuration, latency model, and set of policies, and is one SLURM job with 20 CPU cores and 20\,GB of memory. Within it, every combination of policy, $\lambda_{\text{gen}}$, $\lambda_{\text{lat}}$, and estimated or oracle EI is an independent task of 1,000 Monte Carlo trials for each of the 100 prompts -- between 2,840 and 34,648 tasks per run, depending on how many policies and $\lambda_{\text{lat}}$ values it covers -- and the tasks are spread over the job's cores. Covering the 24 configurations for every experiment in the paper takes about 1,400 such runs. Their median running time is 15 minutes, 90\% finish within 34 minutes, and the longest takes under 4 hours; together they use about 8,800 core-hours. The reported tables and figures are computed from the stored results of these runs, with no further simulation.

\EndAppendixFloats

\end{document}